\documentclass{article}

\PassOptionsToPackage{numbers, compress}{natbib}

\usepackage[preprint]{neurips_2026}

\usepackage[utf8]{inputenc}             %
\usepackage[T1]{fontenc}                %
\usepackage{amsmath}
\usepackage{amssymb}
\usepackage{siunitx}
\usepackage{hyperref}                   %
\hypersetup{
  colorlinks,
  citecolor=blue!60!black,
  linkcolor=black,
  urlcolor=black,
  filecolor=black
}
\usepackage[capitalize]{cleveref}
\usepackage{url}                        %
\usepackage{booktabs}                   %
\usepackage{tabularx}
\usepackage{amsfonts}                   %
\usepackage{nicefrac}                   %
\usepackage{microtype}                  %
\usepackage[dvipsnames,table]{xcolor}   %
\usepackage{pifont}                     %
\usepackage{multirow}
\usepackage{graphicx}
\usepackage{enumitem}
\usepackage{subcaption}
\usepackage{threeparttable}

\usepackage[table]{xcolor}

\newcommand{\tablenoteblock}[1]{%
    \begingroup
        \footnotesize
        \noindent #1\par
    \endgroup
}

\newcommand{\cmark}{\ding{51}}%
\newcommand{\xmark}{\ding{55}}%

\makeatletter
\newcommand*{\@rowstyle}{}
\newcommand*{\rowstyle}[1]{%
  \gdef\@rowstyle{#1}%
  \@rowstyle\ignorespaces%
}
\newcolumntype{=}{>{\gdef\@rowstyle{}}}
\newcolumntype{+}{>{\@rowstyle}}
\makeatother

\title{WayPOP: A Panoramic Open-Set Panoptic Tracking Benchmark}

\author{%
  \textbf{Rohit Mohan\textsuperscript{1,*} \quad
  Maximilian Luz\textsuperscript{1,*} \quad
  Swarnava Chowdhury\textsuperscript{1}} \\
  \textbf{Yao Lu\textsuperscript{1} \quad
  Florian Drews\textsuperscript{2} \quad
  Thomas Nürnberg\textsuperscript{2} \quad
  Yakov Miron\textsuperscript{2,3}} \\
  \textbf{Federico Tombari\textsuperscript{4,5} \quad
  Stefano Gasperini\textsuperscript{4,5,6} \quad
  Abhinav Valada\textsuperscript{1}} \\[0.75em]
  \textsuperscript{1}University of Freiburg \quad
  \textsuperscript{2}Bosch Research \quad
  \textsuperscript{3}University of Haifa \\
  \textsuperscript{4}Technical University of Munich \quad
  \textsuperscript{5}Munich Center for Machine Learning \quad
  \textsuperscript{6}Visualais \\[0.75em]
  \textsuperscript{*}Equal contribution.
}

\begin{document}

\maketitle

\begin{abstract}
  Holistic perception for autonomous agents requires reliably segmenting and associating scene elements across time and camera views, including objects outside a fixed semantic taxonomy.
Existing open-set panoptic benchmarks focus on individual images or views, while panoramic tracking benchmarks assume closed-set taxonomies.
Consequently, these benchmarks do not evaluate whether unknown objects can be consistently segmented and tracked across synchronized multi-camera sequences.
We introduce WayPOP, a panoramic open-set panoptic tracking benchmark built on real Waymo PVPS sequences augmented with diverse out-of-taxonomy 3D assets from Objaverse.
Inserted objects are placed at fixed locations in the 3D scene and rendered using calibrated vehicle-mounted cameras as the ego vehicle and surrounding traffic move, providing temporally and cross-view consistent panoptic annotations and auxiliary LiDAR range observations.
WayPOP contains approximately \num{670} curated assets instantiated as \num{980} unique unknown-object tracks, yielding \num{53452} unknown-object mask annotations across \num{21300} camera images.
Our evaluation protocol defines a hierarchy of tasks from view-independent open-set panoptic segmentation to full panoramic open-set panoptic tracking, using known/unknown PQ and (w)STQ metrics.
We benchmark representative open-set and open-vocabulary segmentation methods with temporal and cross-camera association. %
Our results reveal substantial limitations of current methods in truly open-set scenarios, highlighting robust unknown-object localization and identity association across time and views as major open challenges.
The project page and benchmark are available at \url{https://waypop.cs.uni-freiburg.de}.

\end{abstract}

\section{Introduction}
\label{sec:intro}

\begin{figure}[t]
  \centering
  \includegraphics[width=\linewidth]{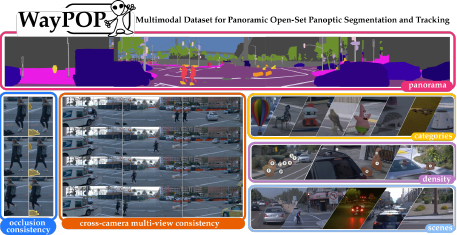}%
  \caption[Overview of the WayPOP Dataset.]{
    \textbf{Overview of the WayPOP Dataset.}
    Real Waymo PVPS driving sequences are augmented with 3D assets rendered consistently across synchronized camera views and time, enabling realistic evaluation of panoramic open-set panoptic segmentation and tracking under viewpoint changes, occlusions, and complex traffic conditions.
    Examples illustrate the diversity of inserted out-of-taxonomy objects, unknown-object densities, and scene contexts.%
  }\label{fig:teaser}%
  \vspace{-1.0em}
\end{figure}

Robots and autonomous agents require coherent scene understanding to operate safely in dynamic environments, with autonomous driving providing a concrete and safety-critical domain for this challenge.
Perception is increasingly moving toward holistic spatiotemporal scene understanding~\cite{luz2026latent}, where \textit{thing} instances are densely segmented and associated over time alongside per-pixel semantic labels for \textit{stuff} classes~\cite{fong2022panoptic, hurtado2020mopt, kim2020video}.
Panoramic multi-camera setups further amplify this challenge: objects frequently leave one camera view and reappear in another, requiring both temporal and cross-camera instance association for unified scene interpretation.
The Waymo Panoramic Video Panoptic Segmentation (Waymo PVPS) benchmark~\cite{mei2022pvps} exemplifies this setting by defining a panoramic panoptic tracking task in which instance identities remain consistent over time and across five synchronized cameras.
However, existing benchmarks are largely restricted to fixed semantic taxonomies and therefore do not evaluate whether perception systems can robustly detect, segment, and associate previously unseen objects across time and views.

This limitation becomes critical in real-world deployment, where autonomous agents inevitably encounter long-tail and out-of-distribution (OOD) objects that fall outside a fixed training taxonomy.
This has motivated a growing body of anomaly, open-set, and OOD perception benchmarks~\cite{pinggera2016lost, chan2021segmentmeifyoucan, hwang2021exemplar, sodano2024open, mohan2024panoptic, bogdoll2024anovox, nekrasov2025spotting}.
However, constructing an open-set benchmark for panoramic video perception introduces additional data-generation challenges.
Real-world anomalous events are rare and difficult to capture at scale, while staged collections~\cite{nekrasov2025spotting} and simulation~\cite{bogdoll2024anovox} trade realism for controllability and scalability.
Image-level synthesis and copy-paste augmentation provide a scalable alternative, but are typically formulated as 2D image-plane operations.
Extending these approaches to panoramic video is nontrivial, as inserted objects must maintain geometrically consistent projections, occlusions, and visible extents across synchronized camera views and time.

To address this gap, we introduce WayPOP, a benchmark for panoramic open-set panoptic segmentation and tracking in autonomous driving (\Cref{fig:teaser}).
WayPOP builds on Waymo PVPS~\cite{mei2022pvps} by augmenting real driving sequences with diverse out-of-taxonomy 3D assets sourced from Objaverse~\cite{deitke2023objaverse}.
Unlike fully synthetic simulators~\cite{mohan2024syn, sekkat2024amodalsynthdrive} or purely 2D copy-paste techniques, WayPOP preserves real-world scene context through geometry-grounded insertion of unknown objects into synchronized multi-camera observations, with auxiliary LiDAR range updates.
Inserted objects remain static in the global 3D scene, while the ego vehicle and surrounding traffic continue to move.
Despite this static placement, our experiments show that it remains challenging to detect, segment, and consistently track unknown objects under viewpoint changes, occlusions, and camera transitions.
Our insertion pipeline is designed to maintain four key properties:
\emph{geometric consistency} (coherent object placement and occlusion relationships),
\emph{photometric consistency} (alignment with scene lighting and conditions),
\emph{multi-view consistency} (coherent projections across synchronized camera views),
and \emph{temporal consistency} (stable object identities over time).
Together, these properties enable controlled evaluation over a highly diverse set of unknown objects while retaining the appearance, occlusion structure, and temporal continuity of real-world driving scenes.

WayPOP contains approximately \num{670} unique Objaverse assets instantiated as \num{980} unknown-object tracks, yielding \num{53452} unknown-object mask annotations across \num{21300} images from five synchronized camera streams with corresponding LiDAR range measurements.
Building on this benchmark, we define \emph{multi-view open-set panoptic tracking} (MV-OSPT), requiring panoptic predictions for known and unknown regions with temporally and cross-view consistent identities, together with a hierarchy of reduced task variants: \emph{view-independent open-set panoptic segmentation} (VI-OSPS) and \emph{frontal-view open-set panoptic tracking} (FV-OSPT).
Finally, we benchmark representative camera-based open-set and open-vocabulary segmentation methods with temporal and cross-camera association, complemented by exploratory multimodal evaluations using LiDAR.

In summary, our main contributions are fivefold:
(1) We introduce WayPOP, to our knowledge the first benchmark for panoramic open-set panoptic tracking in autonomous driving, enabling unknown-object evaluation across synchronized multi-camera video.
(2) We develop a geometry-guided, human-in-the-loop construction pipeline that augments real driving sequences with curated 3D assets, aiming to preserve geometric, photometric, multi-view, and temporal consistency.
(3) We formalize multi-view open-set panoptic tracking (MV-OSPT) and define a hierarchical task suite spanning view-independent segmentation, monocular tracking, and full panoramic tracking.
(4) We provide an open-set panoramic tracking evaluation protocol based on adapted PQ and (w)STQ metrics for separate evaluation of known and unknown regions.
(5) We benchmark representative camera-based open-set and open-vocabulary methods combined with temporal and cross-camera association strategies, revealing substantial limitations in maintaining coherent unknown-object identities across time and views.

\section{Related Work}
\label{sec:related_works}

\begin{table*}[t]
\centering
\caption{\small Comparison of driving benchmarks for anomaly detection, out-of-distribution (OOD) recognition, open-set panoptic segmentation, and panoramic tracking. WayPOP combines unknown-object panoptic labels with temporal and cross-camera identity consistency in real driving sequences.}
\label{tab:dataset_comparison}
\scriptsize
\setlength{\tabcolsep}{1.6pt}
\renewcommand{\arraystretch}{1.18}
\begin{tabularx}{\textwidth}{
@{}
>{\raggedright\arraybackslash}p{1.8cm} %
>{\raggedright\arraybackslash}p{1.7cm} %
>{\centering\arraybackslash}p{0.75cm} %
>{\centering\arraybackslash}p{1.05cm} %
>{\centering\arraybackslash}p{0.67cm}
>{\centering\arraybackslash}p{0.63cm}
>{\centering\arraybackslash}p{0.7cm}
>{\centering\arraybackslash}p{0.75cm} %
>{\centering\arraybackslash}p{0.65cm} %
>{\centering\arraybackslash}p{1.45cm} %
>{\centering\arraybackslash}p{0.65cm} %
>{\raggedright\arraybackslash}X@{}}
\toprule
& & & &
\multicolumn{3}{c}{\textbf{Capabilities}} &
& & \textbf{Size} & & \\
\cmidrule(lr){5-7}
\cmidrule(lr){10-10}
\textbf{Dataset} & \textbf{Variant} & \textbf{LiDAR} & \textbf{Source} &
Temp. & Pano. & Track. & \textbf{\#OOD} & \textbf{Label} &
Test / Val & \textbf{\#Seq.} & \textbf{Data origin} \\
\midrule

\multirow{2}{=}{\textbf{Fishyscapes}~\cite{blum2019fishyscapes}}
& FS Lost\&Found
& \xmark & Staged   & \xmark & \xmark & \xmark & 7  & SM & 275 / 100  & n/a & Lost and Found~\cite{pinggera2016lost} \\
& FS Static
& \xmark & 2D aug.  & \xmark & \xmark & \xmark & 12  & SM & 1,000 / 30  & n/a & Cityscapes~\cite{cordts2016cityscapes} \\
\arrayrulecolor{gray!50}
\specialrule{\lightrulewidth}{0.2ex}{0.2ex}

\multirow{2}{=}{\textbf{CAOS}~\cite{basart2022scaling}}
& StreetHazards
& \xmark & Simulated   & \cmark & \xmark & \xmark & 250  & SM & 1,500 / -     & 150 & \\
& BDD-Anomaly
& \xmark & Val. set  & \xmark & \xmark & \xmark & 3  & SM & 810 / -      & n/a & BDD100K~\cite{yu2020bdd100k} \\
\specialrule{\lightrulewidth}{0.2ex}{0.2ex}

\multirow{2}{=}{\textbf{SMIYC}~\cite{chan2021segmentmeifyoucan}}
& RoadAnomaly21
& \xmark & Web & \xmark & \xmark & \xmark & 26  & SM & 100 / 10   & n/a & \\
& RoadObstacle21
& \xmark & Staged    & \cmark & \xmark & \xmark & 31  & SM & 382 / 30   & 90  & \\
\specialrule{\lightrulewidth}{0.2ex}{0.2ex}

\multirow{2}{=}{\textbf{Wupp.\ OOD}~\cite{maag2022two}}
& SOS
& \xmark & Staged    & \cmark & \xmark & \cmark & 13 & IM & 1,129 / -     & 20  & \\
& CWL
& \xmark & Simulated   & \cmark & \xmark & \cmark & 18 & IM & 1,210 / -     & 26  & \\
\specialrule{\lightrulewidth}{0.2ex}{0.2ex}

\multicolumn{2}{@{}l}{\textbf{Lost\&Found}~\cite{pinggera2016lost}}
& \xmark & Staged    & \cmark & \xmark & \xmark & 37  & SM & 1,068 / -    & 5   & \\
\specialrule{\lightrulewidth}{0.2ex}{0.2ex}

\multicolumn{2}{@{}l}{\textbf{SOD}~\cite{singh2020lidar}}
& \cmark & Staged    & \xmark & \xmark & \xmark & 9   & SM & 460 / 530 & 5   & \\
\specialrule{\lightrulewidth}{0.2ex}{0.2ex}

\multicolumn{2}{@{}l}{\textbf{WD-Pascal}~\cite{bevandic2019simultaneous}}
& \xmark & Simulated   & \xmark & \xmark & \xmark & 6   & SM & - / 70      & n/a & \\
\specialrule{\lightrulewidth}{0.2ex}{0.2ex}

\multicolumn{2}{@{}l}{\textbf{Vistas-NP}~\cite{grcic2020dense}}
& \xmark & Class excl. & \xmark & \xmark & \xmark & 4 & SM & 11,167/-  & n/a & Map.\ Vistas~\cite{neuhold2017mapillary} \\
\specialrule{\lightrulewidth}{0.2ex}{0.2ex}

\multicolumn{2}{@{}l}{\textbf{PANIC}~\cite{sodano2024open}}
& \xmark & Real      & \xmark & \xmark & \xmark & 58  & PM & 679 / 131 & n/a & Cityscapes~\cite{cordts2016cityscapes} \\
\specialrule{\lightrulewidth}{0.2ex}{0.2ex}

\multicolumn{2}{@{}l}{\textbf{AnoVox}~\cite{bogdoll2024anovox}}
& \cmark & Simulated   & \cmark & \cmark & \xmark & 178 & IM & n/a     & 1,228 & CARLA~\cite{dosovitskiy2017carla} \\
\specialrule{\lightrulewidth}{0.2ex}{0.2ex}

\rowcolor{blue!7}
\multicolumn{2}{@{}l}{\textbf{WayPOP (Ours)}}
& \textbf{\cmark} & \textbf{3D aug.} & \textbf{\cmark} & \textbf{\cmark} & \textbf{\cmark} & \textbf{670} & \textbf{PM} & \textbf{19,665 / 1,635} & \textbf{80} & Waymo PVPS~\cite{mei2022pvps} \\

\arrayrulecolor{black}
\bottomrule
\end{tabularx}
\par\smallskip
\parbox{\textwidth}{\scriptsize\raggedright
All datasets provide RGB. Temp.\ = temporal; Pano.\ = panoramic; Track.\ = tracking.
SM/IM/PM = semantic/instance/panoptic masks.}
\end{table*}

\textbf{Closed-set panoptic segmentation and tracking.}
Panoptic segmentation combines semantic segmentation of \textit{stuff} regions with instance segmentation of \textit{thing} objects~\cite{kirillov2019panoptic}, with benchmarks such as Panoptic COCO~\cite{lin2014microsoft}, Cityscapes~\cite{cordts2016cityscapes}, Mapillary Vistas~\cite{neuhold2017mapillary}, and Syn-Mediverse~\cite{mohan2024syn} driving substantial progress~\cite{mohan2021efficientps, cheng2020panoptic, cheng2022masked}.
Panoptic tracking further requires temporally consistent instance identities~\cite{hurtado2020mopt, kim2020video, fong2022panoptic, weber2021step}, while recent panoramic benchmarks such as WildPPS~\cite{jaus2023panoramic} and Waymo PVPS~\cite{mei2022pvps} additionally introduce cross-view consistency across synchronized cameras.
However, these benchmarks remain restricted to closed-set taxonomies and therefore do not evaluate whether previously unseen objects can be segmented and consistently associated across time and views.

\textbf{Open-set and OOD segmentation benchmarks.}
A complementary line of work studies anomaly, out-of-distribution (OOD), and open-set segmentation in driving scenes.
Benchmarks such as Fishyscapes~\cite{blum2019fishyscapes}, Lost and Found~\cite{pinggera2016lost}, and RoadObstacle21 from SegmentMeIfYouCan~\cite{chan2021segmentmeifyoucan} primarily provide semantic or binary anomaly masks, with the latter two addressing world-static road hazards observed from moving vehicles.
SOS~\cite{nekrasov2025spotting} additionally evaluates temporal identity persistence in this setting.
Open-set panoptic benchmarks extend this setting by requiring unknown objects to be segmented as instances while preserving known-class panoptic predictions~\cite{hwang2021exemplar, mohan2024panoptic, sodano2024open, mohan2025open}, but remain predominantly single-view or frame-centric.
JRDB-PanoTrack~\cite{le2024jrdb} is closest in spirit to our work, as it studies panoramic open-world panoptic tracking.
However, it targets human-centric robotic environments rather than on-road autonomous driving with vehicle-mounted cameras and real traffic structures.
As summarized in \cref{tab:dataset_comparison}, existing driving-focused OOD benchmarks provide different combinations of unknown-object annotations, temporal sequences, and panoramic coverage, but none jointly supports open-set panoptic tracking across time and camera views.
WayPOP addresses this missing intersection by combining unknown-object panoptic segmentation with temporally and cross-view consistent identities in real driving sequences.
Furthermore, WayPOP introduces substantially greater OOD object diversity than prior work, enabling more thorough evaluation and better alignment with real-world variability.

\textbf{OOD data generation for autonomous driving.}
Collecting diverse real-world OOD events in active traffic is difficult because anomalous objects are rare and their collection is constrained by safety and logistical considerations.
Existing benchmarks therefore rely on staged capture~\cite{pinggera2016lost, nekrasov2025spotting}, simulation~\cite{bogdoll2024anovox}, validation-set mining~\cite{li2022coda, xia2025openad}, or 2D image-level augmentation~\cite{blum2019fishyscapes, mohan2024panoptic}, each trading off realism, controllability, and scalability.
In particular, image-plane augmentation does not naturally provide geometry-consistent projections and occlusions across synchronized cameras and time.
WayPOP instead inserts curated 3D assets into synchronized multi-camera and LiDAR observations, enabling temporally and cross-view consistent panoramic open-set evaluation in real driving sequences.

\textbf{Open-set segmentation and association baselines.}
Recent open-set~\cite{gasperini2023segmenting, schmidt2025prior2former, mohan2024panoptic, rai2024mask2anomaly, nayal2023rba, lu2026hyp2former} and open-vocabulary panoptic segmentation methods~\cite{yu2023fcclip, jiao2024maftplus, dong2023maskclip, xu2023odise} have improved frame-level recognition and segmentation of novel objects, but are typically applied independently to each image and therefore do not address the temporal and cross-camera consistency required for panoramic panoptic tracking.
Conversely, modern association methods, including box-based trackers~\cite{shim2025focusing, aharon2022bot, zhang2022bytetrack, cao2023observation} and mask propagation approaches~\cite{li2024matching, cheng2023tracking, jiang2026sam2mot, luz2024amodal}, are primarily evaluated in closed-set settings where unknown instances are not explicitly modeled.
WayPOP bridges both lines of work by evaluating open-set panoptic predictions under progressively stronger constraints on temporal and cross-view consistency.

\section{WayPOP Benchmark}
\label{sec:waypop}

\subsection{Task Definition}
\label{ssec:task-definition}

\textbf{Multi-View Open-Set Panoptic Tracking (MV-OSPT).}
We define \emph{multi-view open-set panoptic tracking}, which requires full-scene panoptic predictions for both known classes and out-of-taxonomy objects, with temporally and cross-view consistent instance identities across synchronized camera streams.
At each time step $t \in \{1, \dots, T\}$, the input consists of synchronized multi-camera images $\{ I_t^v \}_{v=1}^V$, and the goal is to produce panoptic predictions $P_t^v$ assigning each pixel a semantic label $y$ and instance identity $k$.
The semantic label space is given by $y \in \mathcal{C}_{\text{known}} \cup \{\text{unknown}\}$, where all out-of-taxonomy objects are mapped to a single \emph{unknown} category.
As in standard panoptic segmentation, \emph{thing} classes, including \emph{unknown}, are segmented into instances with unique identities.
Following Waymo PVPS~\cite{mei2022pvps}, predictions across synchronized camera views are evaluated jointly, requiring consistent instance identities across cameras to form a unified panoramic scene representation.

MV-OSPT imposes two consistency constraints: \emph{temporal consistency}, which requires stable identities over time, and \emph{cross-view consistency}, which requires shared identities across camera views and view transitions.
Together, these constraints require jointly reasoning over space, time, and camera geometry while handling both known and unknown objects.
We derive two simpler task variants by selectively relaxing temporal and cross-view consistency constraints, yielding a hierarchy of tasks:

\textbf{Frontal-View Open-Set Panoptic Tracking (FV-OSPT).}
By removing cross-view consistency while retaining temporal consistency, the task reduces to monocular open-set panoptic tracking in the front-facing camera stream.
Methods must produce temporally consistent panoptic predictions with stable instance identities, including for unknown objects.

\textbf{View-Independent Open-Set Panoptic Segmentation (VI-OSPS).}
Removing both temporal and cross-view consistency reduces the task to per-frame open-set panoptic segmentation across all camera views, with each image processed independently.

All task variants share the same open-set output space and differ only in the consistency constraints imposed on instance identities, forming a hierarchy from independent per-frame segmentation to monocular tracking and full panoramic multi-view tracking.
VI-OSPS enables direct comparison to prior per-frame segmentation methods, FV-OSPT aligns with established monocular tracking benchmarks~\cite{cordts2016cityscapes, geiger2012kitti, yu2020bdd100k}, and MV-OSPT represents the full setting requiring joint reasoning over temporal dynamics, cross-view geometry, and open-set instance segmentation.

\subsection{Evaluation Metrics}
\label{ssec:metrics}

We evaluate segmentation and tracking performance using established metrics adapted to the open-set and panoramic setting.
For VI-OSPS, we report the mean Intersection-over-Union (mIoU) and Panoptic Quality (PQ) for semantic and panoptic segmentation quality, respectively.
For FV-OSPT and MV-OSPT, we report Segmentation and Tracking Quality (STQ)~\cite{weber2021step} and the panoramic weighted STQ (wSTQ) metric from Waymo PVPS~\cite{mei2022pvps}.
To reflect the open-set setting, all metrics are reported separately for known and unknown regions.
For tracking evaluation, Association Quality (AQ) is computed in a class-conditioned manner by restricting ground-truth tracks to the respective subset while leaving predictions unfiltered, enabling evaluation of tracking consistency independently of semantic classification accuracy.

\subsection{Geometry-Guided Data Generation Pipeline}
\label{ssec:pipeline}

\begin{figure}[tb]
    \centering
    \includegraphics[height=3.9cm]{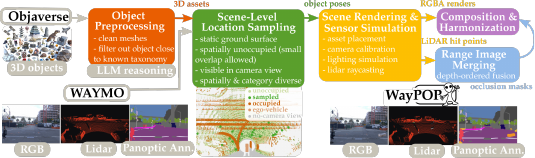}
    \caption[Overview of our geometry-guided data generation pipeline.]{%
        \textbf{Overview of our geometry-guided data generation pipeline.}
        Starting from real-world Waymo sequences and curated 3D assets from Objaverse, objects are inserted at scene-level locations sampled from LiDAR geometry.
        Inserted assets are rendered and composited into camera views and corresponding LiDAR measurements, with updated range images guiding occlusion masks.
        This unified process produces photorealistic RGB images, sensor-consistent LiDAR data, and temporally and cross-view consistent panoptic annotations with inserted objects labeled as \emph{unknown}.%
    }\label{fig:pipeline}%
    \vspace{-1.0em}
\end{figure}

WayPOP is constructed via a geometry-guided, human-in-the-loop pipeline that inserts synthetic 3D assets into synchronized multi-view camera and LiDAR observations from real driving sequences while preserving geometric, photometric, multi-view, and temporal consistency.
The pipeline consists of three stages: (1) asset curation and filtering, (2) geometry-aware scene-level object placement, and (3) physically grounded rendering and sensor integration (\Cref{fig:pipeline}).

\textbf{Source Data and Asset Curation.}
Our pipeline builds on real-world driving sequences from the PVPS subset~\cite{mei2022pvps} of the Waymo Open Dataset~\cite{sun2020waymo}, providing synchronized multi-view camera streams, LiDAR observations, and panoptic annotations.
We adopt the PVPS taxonomy as the set of known classes, treating all inserted objects as out-of-taxonomy.
To populate the unknown object space, we curate 3D assets from Objaverse v1.0~\cite{deitke2023objaverse} and Objaverse-XL~\cite{deitke2023objaverseXL}, which provide complementary trade-offs between structured category annotations and object diversity.
For Objaverse v1.0, we use LVIS labels~\cite{gupta2019lvis} to remove assets that are semantically similar to known driving classes or are implausibly small for driving scenarios.
For Objaverse-XL, which relies on unstructured metadata~\cite{luo2023scalable}, we use Qwen3-14B~\cite{yang2025qwen3} to filter semantically overlapping objects.
Following automatic filtering, we manually remove degenerate or low-quality meshes, correct object scale, and verify driving-scene plausibility.
This process yields a curated pool of approximately \num{670} unique assets for insertion.
We refer to the supplementary material (\Cref{ssec:sup-dataset-curation}) for further details.

\textbf{Scene-Level Insertion Point Sampling.}
Objects are inserted with a fixed pose in the global 3D coordinate frame of each driving sequence rather than independently per frame or camera view.
Inserted assets, therefore, remain static relative to the environment as the ego vehicle moves, and their projections across camera views and time follow directly from the recorded ego poses and sensor calibration.
To identify valid insertion locations, we aggregate LiDAR observations over time and restrict candidates to visible static ground surfaces by excluding dynamic objects and non-ground semantic classes.
While intersections with existing scene elements are generally avoided, we allow a small fraction of overlaps with annotated %
dynamic actors to avoid biasing insertions toward always-empty regions.
Occasional actor-object intersection remains a limitation.
For each candidate location, we estimate a local supporting surface using RANSAC~\cite{fischler1981random} plane fitting and retain only sufficiently large, approximately planar regions.
Candidate locations are prioritized based on visibility and spatial coverage, and a spatially distributed subset is selected in a bird’s-eye-view while maintaining a target distribution over surface types (e.g., road, sidewalk, and other ground).
Objects are assigned from the curated asset pool while enforcing diversity and plausibility constraints.
We avoid repeated object categories within a sequence and, for Objaverse-XL assets, derive coarse semantic categories for this from metadata using Qwen3-14B~\cite{yang2025qwen3}.
Finally, object scale is adapted to the spatial context by assigning smaller objects to nearby locations and larger objects to distant regions, while object pose is sampled randomly.

\textbf{Physically Grounded Rendering and Sensor Integration.}
Given the selected object placements, we generate sensor-consistent observations by rendering inserted assets into all synchronized camera views and integrating them into the corresponding LiDAR observations.
Since placement is defined in the global 3D frame, projections across views and time follow directly from the known sensor calibration.
Inserted assets are rendered using the physically based Blender Cycles engine, accounting for view-dependent geometry and appearance.
Scene lighting is configured to approximate the recording conditions using metadata-derived illumination cues, including sun position and weather information.
Rendered objects are composited into the camera images using depth-based visibility masks derived from the inserted object's geometry, with manual refinement as needed to correct incomplete occlusion relationships.
We further improve visual coherence through photometric harmonization~\cite{guerreiro2023PCTNet}, adapting color and tone of the rendered object to the surrounding image.
Ablations of lighting adaptation, occlusion handling, and harmonization are provided in the supplementary material (\Cref{ssec:sup-dataset-visual}).

To ensure geometric consistency across modalities, we update LiDAR range measurements by raycasting along the original Waymo sensor rays wherever inserted objects occlude existing returns.
Operating directly in the range-image domain preserves the sensor sampling pattern and enforces consistent occlusion relationships between real and inserted content.
Finally, panoptic annotations are generated from the rendering process.
Instance identities are consistent across time and camera views by construction, and all inserted objects are mapped to the \emph{unknown} class.
This unified pipeline ensures joint consistency between RGB observations, LiDAR data, and annotations.
We analyze temporal and geometric consistency under ego motion in the supplementary material (\Cref{ssec:sup-dataset-consistency}).

\subsection{Dataset Structure and Statistics}
\label{ssec:dataset-stats}

\begin{figure}[h]
    \centering
    \captionsetup[subfigure]{font=footnotesize,labelfont=bf,justification=centering}
    \begin{subfigure}[t]{0.24\linewidth}
        \centering
        \includegraphics[width=\linewidth]{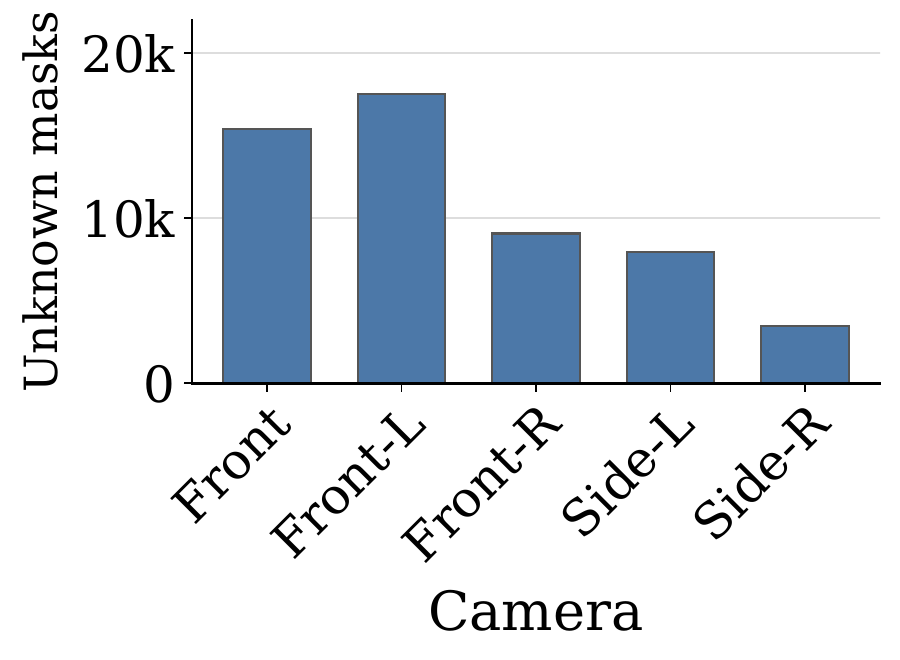}
        \caption{Unknown masks per camera}
        \label{fig:dataset-stats-camera}
    \end{subfigure}\hfill
    \begin{subfigure}[t]{0.24\linewidth}
        \centering
        \includegraphics[width=\linewidth]{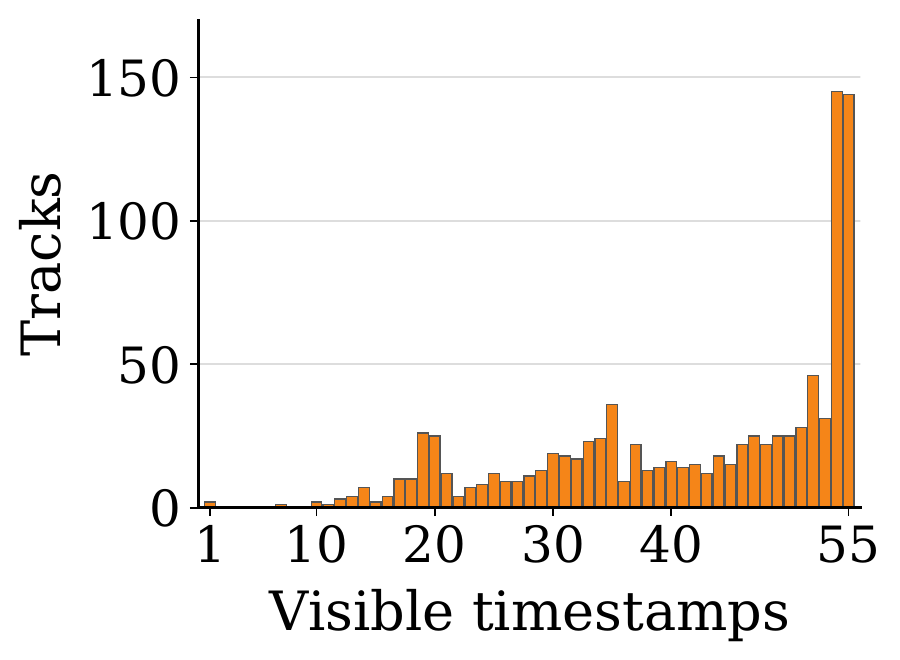}
        \caption{Track duration distribution}
        \label{fig:dataset-stats-duration}
    \end{subfigure}\hfill
    \begin{subfigure}[t]{0.24\linewidth}
        \centering
        \includegraphics[width=\linewidth]{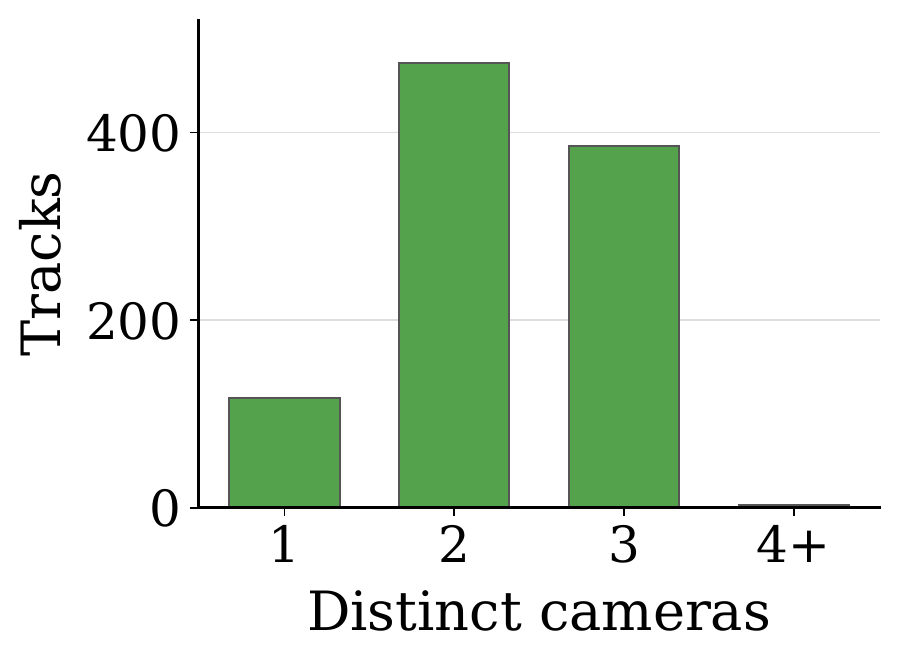}
        \caption{Visible cameras per track}
        \label{fig:dataset-stats-views}
    \end{subfigure}\hfill
    \begin{subfigure}[t]{0.24\linewidth}
        \centering
        \includegraphics[width=\linewidth]{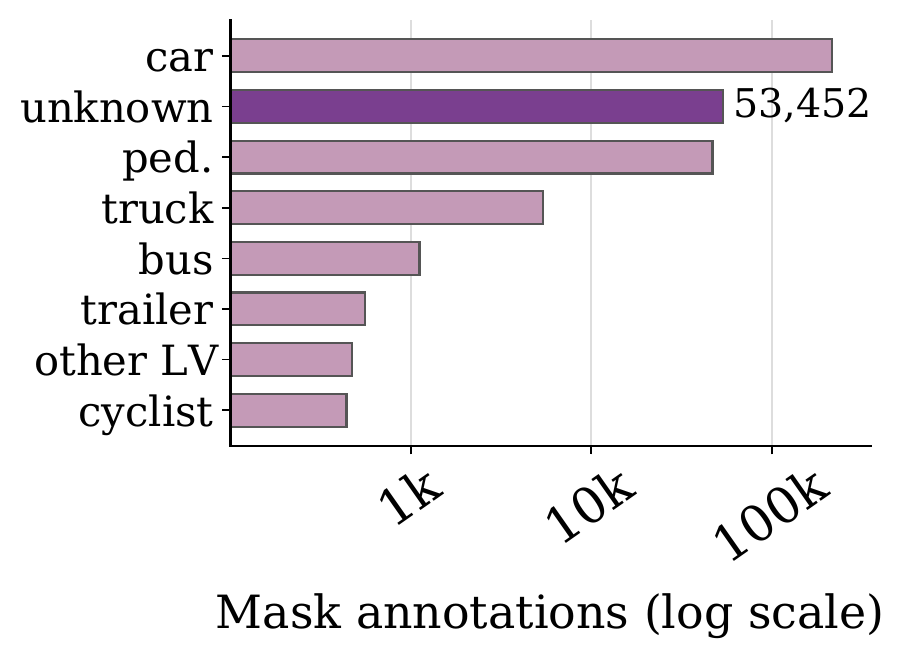}
        \caption{Unknown and known thing masks}
        \label{fig:dataset-stats-known-unknown}
    \end{subfigure}
    \caption{Dataset statistics for WayPOP. We report unknown mask distributions across cameras, temporal and cross-camera track visibility, and mask counts for known versus unknown objects. Mask statistics are per-image; track statistics use resolved visible tracks.
    }
    \label{fig:dataset-stats}
\end{figure}

WayPOP is derived from the \num{20} validation sequences of Waymo PVPS~\cite{mei2022pvps}.
Each sequence is split into two temporal clips, yielding \num{40} base clips, and rendered with two asset-source variants using Objaverse v1.0 and Objaverse-XL.
This results in \num{80} sequence variants across the same real driving scenes, each with a different inserted out-of-taxonomy object. The benchmark consists of three splits: validation (\num{6} V1 clips from \num{3} PVPS sequences), WayPOP-v1 (\num{34} test clips from the remaining \num{17} sequences), and WayPOP-XL (\num{40} test clips from all \num{20} sequences). Unlike Waymo PVPS, which evaluates \num{28} classes, WayPOP retains \num{22} known classes and excludes \texttt{ego\_vehicle}, \texttt{bird}, \texttt{ground\_animal}, \texttt{static}, \texttt{dynamic}, and \texttt{ground}.
Inserted assets are added as instance-aware \emph{unknown} objects.
Each sequence variant provides five synchronized RGB streams, panoptic annotations with temporally consistent identities, and aligned LiDAR insertions, while our benchmark experiments focus on camera-based evaluation.
\Cref{fig:dataset-stats} summarizes the resulting mask distributions and temporal and cross-camera track visibility.
Further asset and split statistics are provided in the supplementary material (\Cref{ssec:sup-dataset-composition}).

\section{Baselines}
\label{sec:baselines}

We evaluate two complementary paradigms for handling out-of-distribution objects: \emph{open-set} and \emph{open-vocabulary} perception.
Open-set methods explicitly model distributional uncertainty, while open-vocabulary methods leverage vision-language models to generalize beyond fixed training taxonomies.
Since these methods operate on a per-image basis, we pair them with temporal and cross-camera association strategies for panoramic tracking evaluation.
Our primary baselines use camera inputs.
Exploratory RGB-LiDAR segmentation experiments are provided in the supplementary material (\Cref{ssec:sup-exp-lidar}).

\subsection{Open-Set and Open-Vocabulary Baselines}

\textbf{Open-Set.}
We evaluate four open-set panoptic segmentation baselines covering complementary mechanisms for identifying unknown objects. U3HS~\cite{gasperini2023segmenting} uses uncertainty-guided unknown prediction, PoDS~\cite{mohan2024panoptic} explicitly models unknown instances, Mask2Anomaly~\cite{rai2024mask2anomaly} provides a mask-centric anomaly formulation, and RbA~\cite{nayal2023rba} serves as a rejection-based baseline. Together, these methods span uncertainty estimation, anomaly segmentation, and rejection-based open-set inference.
We refer to the supplementary material for further details (\Cref{ssec:sup-baselines-os}).

\textbf{Open-Vocabulary.}
We evaluate open-vocabulary panoptic segmentation methods spanning multiple design paradigms, including CLIP-based~\cite{radford2021learning} approaches (FC-CLIP~\cite{yu2023fcclip}, MAFT+~\cite{jiao2024maftplus}, EOV-Seg~\cite{niu2025EOVSeg}), diffusion-based segmentation (ODISE~\cite{xu2023odise}), and prompt-driven segmentation (OpenWorldSAM~\cite{xiao2025owsam}).
All methods are evaluated using publicly available pretrained checkpoints without architectural modifications or additional fine-tuning.
To obtain open-set predictions, all methods use a shared, fixed prompt vocabulary that includes synonyms for known classes and a small set of generic unknown prompts (e.g., \textit{obstacle}, \textit{debris}, \textit{road hazard}).
Since enumerating all possible unknown categories is infeasible, these prompts serve as a proxy for open-set recognition while avoiding overly broad terms such as \textit{object}.
Additional implementation details, the full vocabulary, and a prompt-sensitivity analysis for ODISE are provided in the supplementary material (\Cref{ssec:sup-baselines-ov}). The analysis shows that, across six prompt sets, ODISE's unknown PQ varies substantially, whereas its known PQ remains comparatively stable.

\subsection{Tracking and Panoramic Association Baselines}

The segmentation baselines produce open-set panoptic predictions independently for each camera frame.
To evaluate tracking, we pair each method with temporal and cross-camera association strategies that assign identities to predicted \emph{thing} and \emph{unknown} instances while leaving segmentation outputs unchanged.
We first perform within-camera tracking using box-based~\cite{shim2025focusing, aharon2022bot} and mask-based~\cite{li2024matching, cheng2023tracking, jiang2026sam2mot} methods, producing view-local tracklets.
These tracklets are then projected into the panoramic space~\cite{mei2022pvps} and merged across overlapping cameras using IoU-based matching over short temporal windows.
This associative formulation isolates tracking errors from segmentation quality, enabling separate analysis of temporal drift and cross-camera fragmentation.
Additional analyses with ground-truth masks are provided in the supplementary material (\Cref{ssec:sup-exp-perfect-mask}).

\section{Experiments}
\label{sec:experiment}

We evaluate methods on both the Objaverse v1.0 and XL variants, with additional results provided in the supplementary material.
In the following, we present results across tasks of increasing difficulty, progressively introducing temporal and multi-view consistency constraints.

\subsection{Task1: View-Independent Open-Set Panoptic Segmentation (VI-OSPS)}

We report results in \Cref{tab:vi-osps,tab:vi-osps2} for representative open-vocabulary and open-set methods on WayPOP-v1 and -XL. Current open-vocabulary methods exhibit systematic limitations in truly open-set scenarios: Mask2Former-style~\cite{cheng2022masked} approaches often suppress OOD objects during panoptic inference due to taxonomy-driven foreground selection, while prompt-driven methods such as OpenWorldSAM struggle to localize OOD regions from generic prompts alone. Together, these findings suggest that robust OOD panoptic segmentation requires class-agnostic objectness beyond fixed taxonomies and explicit prompt specification. Further analysis is provided in the supplementary material (\Cref{sssec:sup-baselines-ov-limitations}).
Open-set methods generally achieve higher unknown PQ than the evaluated open-vocabulary approaches, %
suggesting that explicit unknown-object modeling is beneficial in this setting.
RbA and PoDS provide the strongest balance between known- and unknown-class performance, while M2A favors known-class accuracy at the expense of unknown detection and U3HS underperforms on both metrics.
Open-set methods achieve comparable performance across WayPOP-v1 and WayPOP-XL, whereas open-vocabulary methods such as ODISE obtain higher unknown PQ on WayPOP-v1, likely due to recognizable LVIS category names enabling semantic matching.
We therefore focus the main-text tracking evaluation on WayPOP-v1.0.
The supplementary material provides a cross-dataset segmentation comparison (\Cref{ssec:sup-exp-crossdataset}) and additional frontal-view tracking results on WayPOP-XL (\Cref{ssec:sup-exp-fv-ospt}).

\begin{table}[t]
    \caption{%
        VI-OSPS results on the WayPOP Objaverse v1.0 test split for open-vocabulary methods.
    }
    \label{tab:vi-osps}
    \centering
    \begin{center}
\footnotesize
\setlength{\tabcolsep}{4pt}
\sisetup{
  detect-all,
  table-format=2.1
}
\begin{tabular}{ll *{6}{S[table-format=2.1]}}
\toprule
& & \multicolumn{3}{c}{\textbf{Known}} & \multicolumn{3}{c}{\textbf{Unknown}} \\
\cmidrule(lr){3-5} \cmidrule(lr){6-8}
\textbf{Method} & \textbf{Variant}
& {PQ $\uparrow$} & {SQ $\uparrow$} & {RQ $\uparrow$}
& {PQ $\uparrow$} & {SQ $\uparrow$} & {RQ $\uparrow$} \\
\midrule
MAFT+~\cite{jiao2024maftplus} & Large
& 23.6 & 66.4 & 29.8 & 0.1 & 83.6 & 0.1 \\
FC-CLIP~\cite{yu2023fcclip} & ConvNeXt-Large
& \bfseries 27.5 & 61.4 & \bfseries 33.8
& 0.2 & \bfseries 87.4 & 0.2 \\
EOV-Seg~\cite{niu2025EOVSeg} & ConvNeXt-Large
& 25.5 & 64.7 & 31.7 & 0.7 & 85.1 & 0.8 \\
OpenWorldSAM~\cite{xiao2025owsam} & Standard
& 10.8 & 59.7 & 14.2 & 1.0 & 83.8 & 1.2 \\
ODISE~\cite{xu2023odise} & Caption
& 24.1 & \bfseries 67.8 & 29.1
& \bfseries 30.9 & 86.2 & \bfseries 35.8 \\
\bottomrule
\end{tabular}
\end{center}

\end{table}

\begin{table}[t]
    \caption{%
        VI-OSPS results on the WayPOP test split for the Objaverse v1.0 and Objaverse-XL asset variants.
        Evaluating both asset sources enables comparison across different levels of OOD diversity and semantic structure.
    }
    \label{tab:vi-osps2}
    \centering
    \setlength{\tabcolsep}{3pt}
\sisetup{
  detect-all,
  table-format=2.1
}
\small
\begin{tabular}{
l l
*{12}{S[table-format=2.1]}
}
\toprule
& &
\multicolumn{6}{c}{\textbf{WayPOP-v1}} &
\multicolumn{6}{c}{\textbf{WayPOP-XL}} \\
\cmidrule(lr){3-8}
\cmidrule(lr){9-14}
& &
\multicolumn{3}{c}{Known} &
\multicolumn{3}{c}{Unknown} &
\multicolumn{3}{c}{Known} &
\multicolumn{3}{c}{Unknown} \\
\cmidrule(lr){3-5}
\cmidrule(lr){6-8}
\cmidrule(lr){9-11}
\cmidrule(lr){12-14}
&\textbf{Method}
& {PQ $\uparrow$} & {SQ $\uparrow$} & {RQ $\uparrow$}
& {PQ $\uparrow$} & {SQ $\uparrow$} & {RQ $\uparrow$}
& {PQ $\uparrow$} & {SQ $\uparrow$} & {RQ $\uparrow$}
& {PQ $\uparrow$} & {SQ $\uparrow$} & {RQ $\uparrow$} \\
\midrule
{\rotatebox[origin=c]{90}{\textbf{OV}}}&
ODISE~\cite{xu2023odise} (Caption)
    & 24.1 & \bfseries 67.8 & 29.1 & \bfseries 30.9 & 86.2 & \bfseries 35.8
    & 24.4 & \bfseries 66.5 & 29.6 & \bfseries 23.6 & 86.3 & \bfseries 27.4 \\
\arrayrulecolor{gray}\midrule\arrayrulecolor{black}
\multirow{4}{*}{\rotatebox[origin=c]{90}{\textbf{Open Set}}}&
U3HS~\cite{gasperini2023segmenting}
    & 32.8 & 53.6 & 40.6 & 9.3 & 67.5 & 13.8
    & 32.1 & 53.4 & 39.8 & 13.2 & 69.8 & 18.9 \\
&M2A~\cite{rai2024mask2anomaly}
    & \bfseries 46.9 & \bfseries 76.7 & \bfseries 57.0
    & 27.5 & 79.6 & 33.8
    & \bfseries 45.8 & \bfseries 76.9 & \bfseries 55.7
    & 26.1 & 79.5 & 32.3 \\
&RbA~\cite{nayal2023rba}
    & 45.3 & 74.4 & 54.7 & 31.9 & 79.2 & 40.2
    & 44.4 & 74.2 & 54.0 & 29.1 & 80.6 & 36.1 \\
&PoDS~\cite{mohan2024panoptic}
    & 45.1 & 74.4 & 54.6 & \bfseries 34.6 & \bfseries 79.8 & \bfseries 43.4
    & 44.2 & 74.4 & 53.7 & \bfseries 31.1 & \bfseries 81.3 & \bfseries 38.2 \\
\bottomrule
\end{tabular}

\end{table}

\subsection{Task2: Frontal-View Open-Set Panoptic Tracking (FV-OSPT)}

Results for frontal-view tracking on the WayPOP Objaverse v1.0 test split are reported in Table~\ref{tab:results-track-fron}, with additional results for the Objaverse XL test split provided in the supplementary material (\Cref{ssec:sup-exp-fv-ospt}). We observe that open-set panoptic tracking performance is largely driven by the quality of the underlying segmentation, with open-set methods outperforming ODISE and PoDS achieving slightly higher unknown STQ than RbA.
Among trackers, appearance-based methods such as MASA and DEVA achieve higher unknown AQ than the motion-based BoT-SORT, %
which may reflect the importance of appearance cues for associating unknown objects without reliable category priors.
Interestingly, PoDS shows a larger gap between known and unknown STQ than RbA and ODISE, suggesting that its stronger unknown-class segmentation benefits unknown tracking, while known-class tracking remains limited by association errors.

\begin{table}[h]
    \caption{
    FV-OSPT results on the WayPOP Objaverse v1.0 test split.
    We combine image-level open-set and open-vocabulary segmentation with temporal association strategies for frontal-view open-set panoptic tracking.
    }
    \label{tab:results-track-fron}
    \centering
\footnotesize
\setlength{\tabcolsep}{3pt}
\sisetup{
  detect-all,
  table-number-alignment=center,
  round-mode=places,
  round-precision=1
}
\begin{tabular}{
l l
*{6}{S[table-format=2.1]}
}
\toprule
& &
\multicolumn{3}{c}{\textbf{Known}} &
\multicolumn{3}{c}{\textbf{Unknown}} \\
\cmidrule(lr){3-5}
\cmidrule(lr){6-8}
& \textbf{Tracker}
& {STQ $\uparrow$} & {AQ $\uparrow$} & {mIoU $\uparrow$}
& {STQ $\uparrow$} & {AQ $\uparrow$} & {mIoU $\uparrow$} \\
\midrule
\multirow{4}{*}{\rotatebox[origin=c]{90}{\textbf{ODISE}~\cite{xu2023odise}}}
& TrackTrack~\cite{shim2025focusing}
    & 10.5 & 4.6 & 23.8
    & 12.6 & 8.1 & 19.7 \\
& BoT-SORT~\cite{aharon2022bot}
    & \bfseries 10.9 & \bfseries 5.0 & 23.8
    & 11.5 & 6.8 & 19.7 \\
& MASA~\cite{li2024matching}
    & 10.8 & 4.9 & 23.8
    & \bfseries 14.5 & \bfseries 10.7 & 19.7 \\
& DEVA~\cite{cheng2023tracking}
    & \bfseries 10.9 & \bfseries 5.0 & 23.8
    & 14.4 & 10.6 & 19.7 \\
\arrayrulecolor{gray}\midrule\arrayrulecolor{black}
\multirow{4}{*}{\rotatebox[origin=c]{90}{\textbf{PoDS}~\cite{mohan2024panoptic}}}
& TrackTrack~\cite{shim2025focusing}
    & 16.7 & 5.6 & 49.6
    & 26.4 & 16.0 & 43.5 \\
& BoT-SORT~\cite{aharon2022bot}
    & 16.5 & 5.5 & 49.6
    & 25.4 & 14.8 & 43.5 \\
& MASA~\cite{li2024matching}
    & 15.8 & 5.0 & 49.6
    & \bfseries 30.0 & \bfseries 20.7 & 43.5 \\
& DEVA~\cite{cheng2023tracking}
    & \bfseries 19.1 & \bfseries 7.4 & 49.6
    & 29.0 & 19.3 & 43.5 \\
\arrayrulecolor{gray}\midrule\arrayrulecolor{black}
\multirow{4}{*}{\rotatebox[origin=c]{90}{\textbf{RbA}~\cite{nayal2023rba}}}
& TrackTrack~\cite{shim2025focusing}
    & 18.4 & 6.8 & 49.4
    & 26.3 & 18.1 & 38.3 \\
& BoT-SORT~\cite{aharon2022bot}
    & 17.4 & 6.1 & 49.4
    & 25.2 & 16.6 & 38.3 \\
& MASA~\cite{li2024matching}
    & 16.2 & 5.3 & 49.4
    & \bfseries 29.6 & \bfseries 22.9 & 38.3 \\
& DEVA~\cite{cheng2023tracking}
    & \bfseries 20.6 & \bfseries 8.5 & 49.4
    & 28.7 & 21.5 & 38.3 \\
\bottomrule
\end{tabular}

\end{table}

\subsection{Task3: Multi-View Open-Set Panoptic Tracking (MV-OSPT)}

\begin{figure}[t]
  \centering
  \includegraphics[width=0.999\linewidth]{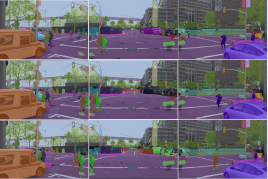}%
  \caption[]{
    \textbf{Qualitative results for MV-OSPT on RbA (TrackTrack)}. The front, front-left and front-right views are presented.
  }\label{fig:qualitative}%
  \vspace{-1.0em}
\end{figure}

Results for panoramic multi-view tracking are reported in \Cref{tab:results-track-mv} with qualitative results for RbA (TrackTrack) presented in Fig.~\ref{fig:qualitative}.
Compared to the frontal-view setting, multi-view tracking substantially increases tracking difficulty due to large viewpoint changes, varying object visibility, and inconsistent appearance across cameras.
Among the evaluated methods, PoDS combined with DEVA achieves the strongest unknown tracking performance, reaching \num{37.1} unknown wSTQ, while RbA with DEVA attains the highest unknown wAQ of \num{25.5}.
Consistent with the frontal-view results, this suggests that stronger OOD sensitivity remains beneficial when identities must persist across time and camera views.
However, substantial association errors remain even with perfect segmentation: given ground-truth masks and classes (wmIoU = \num{100}, marked Oracle), DEVA achieves only \num{43.5} known and \num{61.6} unknown wAQ.

\begin{table}[h]
    \caption{
        MV-OSPT results on the WayPOP Objaverse v1.0 test split.
        Methods combine image-level open-set and open-vocabulary segmentation with temporal and cross-camera association strategies for panoramic open-set panoptic tracking.
        Oracle results use ground-truth masks with the respective tracker to isolate association performance from segmentation errors.
    }
    \label{tab:results-track-mv}
    \centering
\footnotesize
\setlength{\tabcolsep}{3pt}  %

\sisetup{
  detect-all,
  table-number-alignment=center,
  round-mode=places,
  round-precision=1
}

\begin{tabular}{
+l +l
*{6}{+S[table-format=3.1]}
}
\toprule

&
& \multicolumn{3}{c}{\textbf{Known}}
& \multicolumn{3}{c}{\textbf{Unknown}} \\

\cmidrule(lr){3-5}
\cmidrule(lr){6-8}
\textbf{Approach}
& \textbf{Tracker}

& {{wSTQ} $\uparrow$}
& {{wAQ} $\uparrow$}
& {{wmIoU} $\uparrow$}
& {{wSTQ} $\uparrow$}
& {{wAQ} $\uparrow$}
& {{wmIoU} $\uparrow$} \\

\midrule

PoDS~\cite{mohan2024panoptic} & DEVA~\cite{cheng2023tracking}
& 24.2 & 11.4 & 51.3
& 37.1 & 23.9 & 57.6 \\

PoDS~\cite{mohan2024panoptic} & TrackTrack~\cite{shim2025focusing}
& 22.4 & 9.7 & 51.3
& 34.3 & 20.4 & 57.6 \\

RbA~\cite{nayal2023rba} & DEVA~\cite{cheng2023tracking}
& 23.4 & 10.6 & 51.5
& 36.9 & 25.5 & 53.3 \\

RbA~\cite{nayal2023rba} & TrackTrack~\cite{shim2025focusing}
& 22.5 & 9.8 & 51.5
& 34.3 & 22.1 & 53.3 \\

\rowstyle{\color{gray}}
Oracle & DEVA~\cite{cheng2023tracking}
& 65.9 & 43.5 & 100.0
& 78.5 & 61.6 & 100.0 \\

\bottomrule
\end{tabular}

\end{table}

\section{Conclusion}
\label{sec:conclusion}

In this work, we introduced WayPOP, a benchmark for panoramic open-set panoptic tracking in autonomous driving that enables evaluation of unknown-object segmentation and tracking in real driving sequences with long-tail and out-of-taxonomy objects. By augmenting real Waymo PVPS sequences with geometry-guided 3D object insertions, WayPOP supports a hierarchy of tasks ranging from open-set panoptic segmentation to full multi-view open-set tracking. Our experiments show that current open-set and open-vocabulary methods still struggle to robustly localize and maintain consistent unknown-object identities across time and camera views, highlighting panoramic open-set tracking as an important open challenge for autonomous perception. WayPOP nevertheless has several limitations that motivate future work, including imperfect occlusion modeling, static inserted objects, and simplified lighting harmonization. Despite these limitations, we believe WayPOP provides a scalable and realistic foundation for future research on open-world driving perception, dynamic unknown-object insertion, and multimodal panoramic tracking.

\begin{ack}
    This work was funded by the Bosch Research Collaboration on AI-Driven Automated Driving.
    Abhinav Valada was supported by the Deutsche Forschungsgemeinschaft (DFG, German Research Foundation) under Grant 539134284.
    Stefano Gasperini was supported by the German Federal Ministry of Research, Technology and Space (BMFTR) through the Robotics Institute Germany (RIG).
    In addition, the authors would like to thank the German Federal Ministry of Research, Technology and Space and the German federal states (https://go-nhr.de/our-partners) for supporting this work with computing resources as part of the National High-Performance Computing (NHR) joint funding program.

    \paragraph{Licensing.}
WayPOP is not affiliated with, sponsored by, or endorsed by Waymo LLC. WayPOP access requires Waymo Open Dataset registration and acceptance of its non-commercial license.
Objaverse and Objaverse-XL collections use ODC-By v1.0, while individual assets retain their respective licenses.
Original 3D assets are not redistributed; full terms and attribution notices accompany the release.
\end{ack}

\clearpage
\appendix
\pdfbookmark[1]{Supplementary Material}{supplementary}
\section*{Supplementary Material}
\section{Baseline Setup}
\label{sec:sup-baselines}

\subsection{Open-Set Baselines}
\label{ssec:sup-baselines-os}

\subsubsection{Methods}
\label{ssec:sup-baselines-os-methods}

We evaluate four open-set panoptic segmentation methods covering complementary approaches for unknown-object recognition, including uncertainty-guided prediction (U3HS~\cite{gasperini2023segmenting}), explicit unknown-instance modeling (PoDS~\cite{mohan2024panoptic}), mask-based anomaly segmentation (Mask2Anomaly~\cite{rai2024mask2anomaly}), and rejection-based inference (RbA~\cite{nayal2023rba}).
Together, these methods provide a broad basis for assessing how current open-set approaches generalize to unknown objects in WayPOP.

\textbf{U3HS}~\cite{gasperini2023segmenting}.
U3HS combines panoptic segmentation of known classes with uncertainty-based identification of unknown regions.
It separates unknown objects into individual instances by clustering their learned instance-aware embeddings, without requiring unknown-category examples during training.

\textbf{PoDS}~\cite{mohan2024panoptic}.
PoDS jointly predicts known semantic classes and unknown object instances using an OOD contextual module and dual decoders trained with an alignment-mismatch strategy.

\textbf{Mask2Anomaly}~\cite{rai2024mask2anomaly}.
Mask2Anomaly combines mask classification with global masked attention, mask contrastive learning, and mask refinement to identify unknown objects while reducing false positives.

\textbf{RbA}~\cite{nayal2023rba}.
RbA builds on mask classification and identifies unknown regions using an outlier score that measures rejection by all known classes.

We additionally evaluated Prior2Former~\cite{schmidt2025prior2former}, but, despite extensive experimentation and hyperparameter tuning, could not obtain meaningful results on WayPOP.
We therefore omit it from the quantitative comparison.

\subsubsection{Implementation Details}
\label{ssec:sup-baselines-os-parameters}

All four baselines are trained on the Waymo PVPS training set using a batch size of \num{16}, $640\times960$ crops ($H\times W$), and random horizontal flipping.
U3HS, Mask2Anomaly, and RbA use ImageNet-pretrained ResNet-50 backbones.
Mask2Anomaly and RbA share an initial \num{60000}-iteration training setup with AdamW, an initial learning rate of \num{1e-4} (\num{1e-5} for the backbone), weight decay of \num{0.05}, and polynomial learning-rate decay with power \num{0.9}.
Both use random resizing with scale factors from \num{0.5} to \num{2.0}, random cropping, and SSD color augmentation.
They also incorporate COCO 2017 outliers~\cite{lin2014microsoft} with probability \num{0.2} in their respective training stages.
At inference, both use query confidence and overlap thresholds of \num{0.8}.

\textbf{U3HS}~\cite{gasperini2023segmenting}
We use an output stride of \num{16} and \num{8}-dimensional embeddings.
We train for \num{120} epochs using Adam with initial learning rates of \num{1e-3} for the model and \num{5e-5} for the backbone, a per-epoch decay factor of \num{0.98}, and no weight decay.
Data augmentation includes random zoom factors from \num{1.0} to \num{2.0} and SSD color augmentation.
The semantic Dirichlet NLL, center MSE, prototype-instance, and discriminative losses have unit weights.
At inference, unknown instances are obtained using DBSCAN with $\epsilon=\num{0.5}$, discarding instances smaller than \num{200} pixels.
Center-based grouping uses a threshold of \num{0.1}.

\textbf{PoDS}~\cite{mohan2024panoptic}.
We initialize the HRNet-W32 backbone with pretrained weights and train the full model for \num{60000} iterations on \num{22} known classes.
We then train the OOD branches and contextual/dynamic modules for \num{40000} additional iterations, keeping the backbone and base branches frozen.
Both stages use Adam with an initial learning rate of \num{1e-3} and polynomial learning-rate decay with power \num{0.9}.
For OOD training, we apply COCO 2017 copy-paste augmentation~\cite{lin2014microsoft} with probability \num{0.5}, excluding categories overlapping with known Waymo classes.
At inference, we use known and unknown center thresholds of \num{0.5} and \num{0.03}, respectively.

\textbf{Mask2Anomaly}~\cite{rai2024mask2anomaly}
We train the full Mask2Former-based model with \num{100} queries using the shared first-stage setup.
COCO 2017 outliers are inserted during training, with pasted regions ignored in the panoptic supervision. The anomaly score is one minus the clamped maximum known-class response.
Unknown instances are extracted from pixels with anomaly scores of at least \num{0.8} using eight-connected components with a minimum area of \num{1024} pixels.

\textbf{RbA}~\cite{nayal2023rba}
Starting from the final checkpoint of the shared first-stage setup, we fine-tune only the class and mask prediction heads for \num{5000} additional iterations.
The second stage uses the same AdamW settings, COCO 2017 outliers are inserted during this stage.
The squared-hinge RbA loss has weight \num{1.0}, with inlier and outlier margins of \num{-1.0} and \num{-0.1}, respectively. At inference, we use an RbA rejection threshold of \num{-0.2} and a minimum unknown-component area of \num{200} pixels (\Cref{tab:rba-sensitivity}).
We apply $3\times3$ morphological opening followed by closing to the unknown masks, then extract four-connected components to obtain unknown-object instances.

\subsubsection{Parameter Sensitivity}
\label{ssec:sup-baselines-os-sensitivity}

We examine the sensitivity of RbA~\cite{nayal2023rba} to two inference parameters: the unknown-mask threshold and the minimum connected-component area.
\Cref{tab:rba-sensitivity} reports unknown PQ on the validation split across nine parameter combinations. Performance ranges from \num{18.73} to \num{22.67} PQ, with the highest value obtained at a threshold of \num{-0.20} and a minimum area of \num{200}.

\begin{table}[h]
    \centering
    \begin{threeparttable}
        \caption{%
            RbA unknown PQ on the validation split for different unknown-mask thresholds and minimum connected-component areas.
        }
        \label{tab:rba-sensitivity}
        \small
        \sisetup{
  detect-all,
  table-format=2.2
}
\begin{tabular*}{0.8\linewidth}{@{\extracolsep{\fill}}l*{3}{S}}
    \toprule
    & \multicolumn{3}{c}{\textbf{Minimum component area}} \\
    \cmidrule(lr){2-4}
    \textbf{Threshold} & {100} & {200} & {500} \\
    \midrule
    \num{-0.30} & 18.73 & 20.51 & 21.55 \\
    \num{-0.20} & 21.09 & \bfseries 22.67 & 21.40 \\
    \num{-0.10} & 20.41 & 21.25 & 21.81 \\
    \bottomrule
\end{tabular*}

        \tablenoteblock{%
            The best evaluated setting is highlighted in bold.
        }
    \end{threeparttable}
\end{table}

\subsection{Open-Vocabulary Baselines}
\label{ssec:sup-baselines-ov}

\subsubsection{Methods}
\label{sssec:sup-baselines-ov-methods}

We evaluate a diverse set of open-vocabulary panoptic segmentation methods spanning multiple design paradigms, including CLIP-based~\cite{radford2021learning} approaches (FC-CLIP~\cite{yu2023fcclip}, MAFT+~\cite{jiao2024maftplus}, EOV-Seg~\cite{niu2025EOVSeg}), diffusion-based segmentation (ODISE~\cite{xu2023odise}), and prompt-driven segmentation (OpenWorldSAM~\cite{xiao2025owsam}).
Together, these methods provide representative coverage of established and emerging approaches to open-vocabulary panoptic segmentation.

\textbf{FC-CLIP}~\cite{yu2023fcclip}.
FC-CLIP uses a frozen convolutional CLIP backbone for dense prediction, directly leveraging pretrained vision-language representations without task-specific fine-tuning.
Throughout our evaluations, we use the ConvNeXt-Large~\cite{liu2022convnet} variant.

\textbf{MAFT+}~\cite{jiao2024maftplus}.
MAFT+ improves vision-language alignment through collaborative vision-text representation learning and mask-aware fine-tuning of region-level features.
We use the Large variant with the released panoptic segmentation checkpoint.

\textbf{EOV-Seg}~\cite{niu2025EOVSeg}.
EOV-Seg is an efficient open-vocabulary panoptic segmentation method that unifies mask prediction and classification in a streamlined architecture optimized for scalable inference.
We use the ConvNeXt-Large variant with the released panoptic segmentation checkpoint.

\textbf{ODISE}~\cite{xu2023odise}.
ODISE performs open-vocabulary panoptic segmentation using representations derived from pretrained text-to-image diffusion models.
We use the caption-supervised checkpoint.
The label-supervised variant exhibits the unknown-object suppression discussed in \Cref{sssec:sup-baselines-ov-limitations}.

\textbf{OpenWorldSAM}~\cite{xiao2025owsam}.
OpenWorldSAM extends SAM2~\cite{ravi2024sam2} with text-driven mask generation, combining promptable segmentation with language-based object specification.
We use the standard segmentation checkpoint.

All methods are evaluated using publicly available pretrained checkpoints, without architectural modifications or additional fine-tuning on our dataset, reflecting the open-vocabulary setting in which models are expected to generalize without task-specific training.

\subsubsection{Prompt Set}
\label{sssec:sup-baselines-ov-prompts}

We define a fixed set of textual prompts, selected on the validation split and shared across all open-vocabulary methods during test-set evaluation, to ensure a fair, controlled comparison among different architectures and inference paradigms.
Each class is represented by a set of synonyms to improve the robustness of language-based matching.
Since it is not feasible to enumerate all possible out-of-distribution object categories, we additionally include a small set of generic prompts intended to capture common real-world OOD objects (e.g., \textit{debris}, \textit{road hazard}, \textit{foreign object}).
These prompts are chosen to provide meaningful semantic grounding while avoiding overly broad terms (e.g., \textit{object}) that could lead to trivial assignments.
The first term (bold) corresponds to the canonical class name.

\begin{itemize}[noitemsep]
\item \textbf{Car}: car, cars, automobile, minivan, sedan, SUV, hatchback, coupe
\item \textbf{Truck}: truck, trucks, motortruck, pickup truck, cargo van, delivery van, sprinter van
\item \textbf{Bus}: bus, buses, autobus, motorbus, motorcoach, omnibus
\item \textbf{Other large vehicle}: other large vehicle, large vehicle, mobile crane, crane truck, construction vehicle, excavator, bulldozer, camper van, motorhome
\item \textbf{Bicycle}: bicycle, bicycles, bike, bikes
\item \textbf{Motorcycle}: motorcycle, motorcycles, motorbike, scooter, moped, motor scooter
\item \textbf{Trailer}: trailer, semi-trailer, caravan
\item \textbf{Pedestrian}: pedestrian, person, people, walker, child, girl, boy, woman, man, children, girls, boys, women, men, lady, guy, ladies, guys, clothes
\item \textbf{Cyclist}: cyclist, bicyclist, bicyclists, bike rider
\item \textbf{Motorcyclist}: motorcyclist, motorcyclists, motorcycle rider
\item \textbf{Construction cone}: construction cone, traffic cone, road cone, pylon, construction cone pole
\item \textbf{Pole}: pole, poles, post, utility pole, lamp post, light pole, street lamp, street light
\item \textbf{Stroller}: stroller, pram, wheelchair, shopping cart, luggage, trolley
\item \textbf{Sign}: sign, signs, signboard, road sign, traffic sign, street sign, street name sign, road name sign, warning sign, construction sign
\item \textbf{Traffic light}: traffic light, traffic lights, traffic signal
\item \textbf{Building}: building, buildings, edifice, edifices, house, shop, store, storefront, warehouse, garage
\item \textbf{Road}: road, roads, street, streets, route, highway, asphalt, tarmac
\item \textbf{Lane marker}: lane marker, lane marking, lane line
\item \textbf{Road marker}: road marker, road marking, pavement marking
\item \textbf{Sidewalk}: sidewalk, pavement, footpath, walkway
\item \textbf{Vegetation}: vegetation, tree, trees, plant, plants, bushes, foliage, flora, grass, lawn, shrub, hedge
\item \textbf{Sky}: sky, clouds
\item \textbf{Unknown/OOD}: foreign object, road debris, debris, obstacle, fallen object, stray object, loose object, lost cargo, spilled cargo, road hazard
\end{itemize}

\subsubsection{Prompt Sensitivity}
\label{sssec:sup-baselines-ov-sensitivity}

We evaluate sensitivity of ODISE~\cite{xu2023odise} to unknown-object prompts on WayPOP-v1 by comparing six prompt sets, ranging from generic object descriptions to road-hazard terms and our full benchmark vocabulary. Table~\ref{tab:odise-prompt-sensitivity} reports the resulting known and unknown PQ.
Unknown PQ varies substantially across prompt sets, while known PQ remains comparatively stable.
The full vocabulary used in our benchmark achieves the highest unknown PQ among the evaluated alternatives, highlighting the influence of prompt choice on open-set performance.

\begin{table}[h]
    \centering
    \begin{threeparttable}
        \caption{Sensitivity of ODISE to unknown-object prompts on the WayPOP-v1 test split.}
        \label{tab:odise-prompt-sensitivity}
        \small
        \sisetup{detect-all, table-format=2.1}
        \begin{tabular}{@{}p{0.64\linewidth}SS@{}}
            \toprule
            \textbf{Unknown prompts} & {\textbf{Known PQ $\uparrow$}} & {\textbf{Unknown PQ $\uparrow$}} \\
            \midrule
            unknown object & 24.1 & 7.7 \\
            foreign object & 24.2 & 12.4 \\
            object, thing, item, entity & 24.0 & 17.7 \\
            unknown object, unseen object, novel object, foreign object
                & 24.3 & 17.4 \\
            road hazard, road debris, fallen object, obstacle
                & 24.4 & 23.3 \\
            Ours (full vocabulary in Section~\ref{sssec:sup-baselines-ov-prompts})
                & \bfseries 24.1 & \bfseries 30.9 \\
            \bottomrule
        \end{tabular}
    \end{threeparttable}
\end{table}

\subsubsection{Failure Modes and Limitations}
\label{sssec:sup-baselines-ov-limitations}

We observe consistent limitations of current open-vocabulary methods when applied to out-of-distribution (OOD) objects: mask-first approaches (e.g., FC-CLIP~\cite{yu2023fcclip}, MAFT+~\cite{jiao2024maftplus}, ODISE~\cite{xu2023odise}) often localize OOD regions in semantic segmentation but fail to produce corresponding segments in the panoptic setting, while prompt-driven methods such as OpenWorldSAM~\cite{xiao2025owsam} fail to reliably capture OOD objects in both semantic and panoptic outputs.

This behavior stems from how Mask2Former-style~\cite{cheng2022masked} methods couple class prediction with panoptic inference.
Their classification head is trained on a fixed set of in-distribution categories and a void class that absorbs everything outside this taxonomy, effectively deciding whether a region belongs to one of the known classes rather than whether it is foreground.
Open-vocabulary prompts do not influence this foreground selection, and OOD regions are therefore typically assigned high void probability.
In panoptic inference, this void probability is used to filter and remove queries before mask aggregation.
Since open-vocabulary classification is applied only after this foreground selection, queries covering OOD regions are often removed beforehand and cannot be assigned to OOD prompts.
Open-vocabulary reasoning is thus restricted to regions already considered foreground under the model’s in-distribution classifier.
In contrast, semantic inference discards the void class and aggregates predictions across all open-vocabulary prompts, forcing each pixel to be assigned to the most likely foreground category.
This allows generic OOD prompts (e.g., \textit{debris}, \textit{obstacle}) to capture OOD regions despite low confidence.

Notably, ODISE trained with caption-based supervision performs substantially better on OOD objects than label-based variants.
Unlike label supervision, which ties foreground selection to a fixed taxonomy, caption-based training induces a class-agnostic objectness signal that is not restricted to predefined categories.
This allows regions corresponding to arbitrary objects to be retained as foreground, increasing the likelihood that OOD regions survive the filtering stage and can be assigned open-vocabulary labels.

In contrast, prompt-driven methods such as OpenWorldSAM exhibit a different limitation.
Since segmentation is conditioned directly on textual prompts, OOD objects can only be detected if they are well captured by the provided prompts.
Generic OOD prompts (e.g., \textit{debris}, \textit{foreign object}) often provide weak or ambiguous grounding, leading to missing or low-confidence masks in both semantic and panoptic outputs.

Overall, these results indicate that current open-vocabulary methods remain fundamentally limited in truly open-set scenarios.
Mask-first approaches suppress OOD objects due to taxonomy-driven foreground selection, while prompt-driven methods fail to localize them without sufficiently specific prompts.
Together, this suggests that robust OOD panoptic segmentation requires a notion of class-agnostic objectness that is decoupled from fixed training taxonomies and does not rely on explicit prompt specification.

\section{Dataset Construction and Validation}
\label{sec:sup-dataset}

\subsection{Asset Curation and Filtering}
\label{ssec:sup-dataset-curation}

Asset curation aims to select valid 3D objects outside the known Waymo PVPS taxonomy that are suitable for insertion into driving scenes.
While Objaverse v1.0 assets are filtered using LVIS category labels, Objaverse-XL requires metadata-based screening due to the absence of comparable structured annotations.
We detail this screening and the subsequent visual verification below.

\paragraph{Metadata-based screening.}

We use Qwen3-14B~\cite{yang2025qwen3} to screen Objaverse-XL assets based on their available metadata.
The prompt includes the known Waymo PVPS categories, Qwen-generated synonyms, and examples of assets to retain or exclude.
This supports identifying semantic overlap beyond exact keyword matches, including vehicle brands and subtypes.
For each asset, the model predicts a primary identity, semantic and quality flags, known-class overlap indicators, and size and insertability estimates.
It also returns a \texttt{keep}/\texttt{review}/\texttt{drop} recommendation, confidence, and a brief rationale.
These outputs are intended to narrow the candidate pool for subsequent visual verification rather than determine final asset selection.

\paragraph{Visual Verification and Grouping.}

Retained candidates are rendered from three canonical viewpoints and grouped into coarse semantic families using a second metadata-based Qwen pass.
These groupings facilitate browsing through a verification interface, where we manually inspect candidates for visual identity, mesh quality, and semantic overlap with known classes.
Candidates with unsuitable or degenerate geometry, poor mesh quality, or known-class overlap are discarded.
Final asset selection therefore remains human.

\paragraph{Limitations and Alternatives.}
Metadata-based screening depends on descriptions that may be incomplete or misleading.
Since Qwen does not inspect mesh geometry, its predicted identities and size estimates can disagree with the rendered assets.
To address these limitations, we investigated VLM-based captioning of three rendered viewpoints per asset, but found the generated captions insufficiently reliable for filtering.
In addition, this alternative also requires rendering every candidate before screening, increasing the cost of curation.
Fully manual filtering, on the other hand, is impractical at the scale of Objaverse-XL.
We therefore combine metadata-based screening with manual visual verification.

\subsection{Dataset Composition and Statistics}
\label{ssec:sup-dataset-composition}

\subsubsection{Dataset Structure and Splits}
\label{sssec:sup-dataset-composition-splits}

WayPOP is constructed from \num{20} Waymo PVPS validation sequences, each divided into two temporal clips and augmented with objects from Objaverse v1.0 and Objaverse-XL.
This yields \num{40} base clips and \num{80} augmented clips.
The benchmark consists of three splits: validation (\num{6} V1 clips from \num{3} sequences), WayPOP-v1 (\num{34} test clips from the remaining \num{17} sequences), and WayPOP-XL (\num{40} test clips from all \num{20} sequences).
The V1 validation and test splits therefore have no overlapping source sequences, while WayPOP-XL also includes the validation driving sequences with different inserted objects.
\Cref{tab:split-statistics} summarizes the resulting camera-image, unknown-mask, and unknown-track counts.

\begin{table}[h]
    \centering
    \begin{threeparttable}
        \caption{Dataset statistics for the WayPOP validation and test splits.}
        \label{tab:split-statistics}
        \small
        \sisetup{
  detect-all
}
\begin{tabular}{llS[table-format=5.0]S[table-format=5.0]S[table-format=3.0]}
    \toprule
    \textbf{Split} & \textbf{Partition} & {\textbf{Camera Images}}
        & {\textbf{Unknown Masks}} & {\textbf{Unknown Tracks}} \\
    \midrule

    Validation & Val  & 1635  & 2559  & 48  \\
    WayPOP-v1  & Test & 9015  & 22766 & 439 \\
    WayPOP-XL  & Test & 10650 & 28127 & 493 \\

    \arrayrulecolor{gray}\midrule\arrayrulecolor{black}

    \textbf{Total} & & \bfseries 21300
        & \bfseries 53452 & \bfseries 980 \\
    \bottomrule
\end{tabular}

        \tablenoteblock{%
            Camera images count individual camera observations,
            unknown masks count per-image instance annotations,
            and unknown tracks count distinct object identities
            associated across time and camera views.
        }
    \end{threeparttable}
\end{table}

\subsubsection{Asset Composition}
\label{sssec:sup-dataset-composition-assets}

Our curated object pool comprises approximately \num{670} unique assets: \num{270} from Objaverse v1.0 and \num{400} from Objaverse-XL.
Objaverse v1.0 examples include alligators, cupboards, guitars, kegs, drones, and trampolines.
For Objaverse-XL, identities inferred from metadata include geometric sculptures, four-legged robots, dragons, fantasy aircraft, wooden crates, and asteroids.
Note that these descriptions are approximate and may not always match the rendered geometry.

\begin{figure}[htbp]
    \centering
    \includegraphics[width=0.6\linewidth]{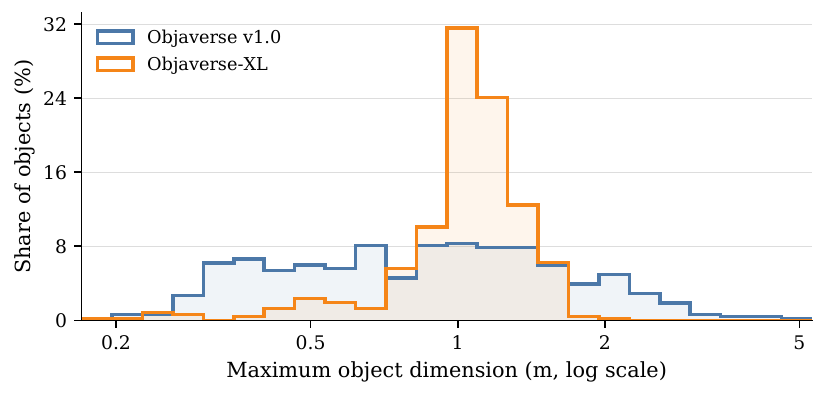}
    \caption{Distributions of maximum object dimensions by asset source, estimated from reconstructed object point clouds. Histograms are normalized separately for each source and use shared logarithmically spaced bins on a logarithmic horizontal axis.}
    \label{fig:sup-object-size-distribution}
\end{figure}

Maximum object dimensions are centered near \qty{1}{\metre} in both sources (\Cref{fig:sup-object-size-distribution}).
For Objaverse v1.0, the median maximum dimension is \qty{0.87}{\metre} (interquartile range: \qtyrange{0.51}{1.36}{\metre}; 10th--90th percentiles: \qtyrange{0.35}{2.04}{\metre}), compared with \qty{1.06}{\metre} for Objaverse-XL (\qtyrange{0.95}{1.20}{\metre} and \qtyrange{0.75}{1.40}{\metre}, respectively).
Objaverse v1.0 exhibits a broader size distribution and a longer tail towards larger objects, while Objaverse-XL is more concentrated around \qty{1}{\metre}.

\subsection{Visual Integration and Pipeline Ablations}
\label{ssec:sup-dataset-visual}

To assess the contribution of the visual integration stages described in \Cref{ssec:pipeline}, we compare four cumulative pipeline variants: rendering without lighting adaptation or occlusion handling, lighting adaptation alone, lighting with occlusion handling, and the full pipeline with additional photometric harmonization.
\Cref{tab:pipeline-fid-kid} compares the resulting images with the original Waymo observations using Fr\'echet Inception Distance (FID)
and Kernel Inception Distance (KID), while \Cref{tab:pipeline_segmentation} reports the corresponding VI-OSPS performance of RbA~\cite{nayal2023rba} and Mask2Anomaly~\cite{rai2024mask2anomaly} on WayPOP-v1.

\begin{table}[h]
    \centering
    \begin{threeparttable}
        \caption{%
            Visual distribution comparison between pipeline variants and original Waymo images on WayPOP-v1.
        }
        \label{tab:pipeline-fid-kid}
        \small
        \sisetup{
  detect-all,
  separate-uncertainty
}
\begin{tabular}{lS[table-format=2.3]S[table-format=1.3(3)]}
    \toprule
    \textbf{Pipeline Variant} & {\textbf{FID} $\downarrow$}
        & {\textbf{KID}$\times 10^3$ $\downarrow$} \\
    \midrule

    No lighting adaptation, no occlusion
        & 14.762 & 0.544 \pm 0.115 \\
    Lighting adaptation, no occlusion
        & 11.804 & 0.206 \pm 0.056 \\
    Lighting + occlusion
        & 11.676 & 0.181 \pm 0.061 \\

    \arrayrulecolor{gray}
    \midrule
    \arrayrulecolor{black}

    Full: lighting + occlusion + harmonization
        & \bfseries 9.745 & \bfseries 0.057 \pm 0.056 \\
    \bottomrule
\end{tabular}

        \tablenoteblock{%
            Variants represent cumulative pipeline additions.
            Lower FID and KID as the mean ± standard deviation indicate smaller distribution differences.
            Best values are highlighted in bold.
        }
    \end{threeparttable}
\end{table}

\begin{table}[h]
    \centering
    \begin{threeparttable}
        \caption{%
            VI-OSPS performance across cumulative pipeline variants on a development version of WayPOP-v1.
        }
        \label{tab:pipeline_segmentation}
        \small
        \setlength{\tabcolsep}{4pt}
\sisetup{
  detect-all,
  table-format=2.1
}
\begin{tabular}{ll*{6}{S}}
    \toprule
    & & \multicolumn{3}{c}{\textbf{Known}}
      & \multicolumn{3}{c}{\textbf{Unknown}} \\
    \cmidrule(lr){3-5}\cmidrule(lr){6-8}
    \textbf{Method} & \textbf{Pipeline Variant}
        & {PQ $\uparrow$} & {SQ $\uparrow$} & {RQ $\uparrow$}
        & {PQ $\uparrow$} & {SQ $\uparrow$} & {RQ $\uparrow$} \\
    \midrule

    RbA~\cite{nayal2023rba} & No lighting adaptation, no occlusion
        & 44.1 & 69.8 & 53.7 & 38.0 & 83.4 & 45.6 \\
    & Lighting adaptation, no occlusion
        & 44.1 & 69.9 & 53.6 & 33.3 & 80.2 & 41.6 \\
    & Lighting + occlusion
        & 44.0 & 69.5 & 53.5 & 32.4 & 79.8 & 40.7 \\
    & Full pipeline
        & 43.7 & 69.7 & 53.0 & 29.9 & 79.1 & 37.8 \\

    \arrayrulecolor{gray}
    \midrule
    \arrayrulecolor{black}

    Mask2Anomaly~\cite{rai2024mask2anomaly} & No lighting adaptation, no occlusion
        & 47.4 & 77.3 & 57.5 & 38.4 & 81.4 & 47.2 \\
    & Lighting adaptation, no occlusion
        & 47.1 & 76.9 & 57.1 & 29.4 & 79.5 & 37.0 \\
    & Lighting + occlusion
        & 47.0 & 76.9 & 56.9 & 27.8 & 79.7 & 34.9 \\
    & Full pipeline
        & 46.8 & 76.8 & 56.8 & 24.3 & 78.7 & 30.9 \\

    \bottomrule
\end{tabular}

        \tablenoteblock{%
            The full pipeline includes lighting adaptation, occlusion handling, and harmonization.
        }
    \end{threeparttable}
\end{table}

Adding lighting adaptation, occlusion handling, and harmonization progressively reduces FID and KID, indicating a smaller distribution gap to the original Waymo images.
At the same time, unknown PQ decreases from \num{38.0} to \num{29.9} for RbA and from \num{38.4} to \num{24.3} for Mask2Anomaly, while known PQ remains comparatively stable.
The decrease in unknown performance is more pronounced in RQ than SQ, suggesting that the integration stages primarily affect unknown-object recognition rather than mask alignment.
Together, these trends are consistent with a reduction in visual artifacts that could otherwise provide shortcuts for identifying inserted objects as unknown, making them more challenging for the evaluated segmentation methods.

Nevertheless, these aggregate comparisons do not establish artifact-free integration or identify the cues exploited by the segmentation methods.
Inserted objects occupy only a small fraction of each image, limiting the sensitivity of whole-image FID and KID to local inconsistencies.
Metadata-based lighting and photometric harmonization also cannot fully reproduce scene-specific shadows, indirect illumination, material appearance, or camera response, leaving a residual photometric gap.

\subsection{Temporal and Geometric Consistency}
\label{ssec:sup-dataset-consistency}

Since WayPOP inserts world-static objects, their geometry should remain consistent under the recorded ego motion, with residual measurement errors comparable to those of native static Waymo structures.
To assess this, we transform mask-associated LiDAR points between consecutive frames using the recorded SE(3) ego poses and measure residual point-set velocity and surface mismatch.
We compute the same statistics for traffic signs and independently verified stationary vehicles as a reference for measurement variability.

The results in \Cref{tab:temporal-consistency} show broadly comparable residuals for inserted and native static objects across ego-vehicle speed ranges, supporting the temporal and geometric consistency of our insertions.

\begin{table}[h]
    \centering
    \begin{threeparttable}
        \caption{%
            Temporal and geometric consistency under ego motion.
        }
        \label{tab:temporal-consistency}
        \small
        \sisetup{
  detect-all,
  table-format=1.3
}
\begin{tabular}{l*{4}{S@{\,/\,}S}}
    \toprule
    & \multicolumn{4}{c}{\textbf{WayPOP-v1.0}}
    & \multicolumn{4}{c}{\textbf{WayPOP-XL}} \\
    \cmidrule(lr){2-5}\cmidrule(lr){6-9}

    \textbf{Ego speed} & \multicolumn{2}{c}{Inserted} & \multicolumn{2}{c}{\begin{tabular}[b]{@{}c@{}}Native\\reference\end{tabular}}
              & \multicolumn{2}{c}{Inserted} & \multicolumn{2}{c}{\begin{tabular}[b]{@{}c@{}}Native\\reference\end{tabular}} \\
    \midrule

    \qtyrange{1}{5}{\metre\per\second}  & 0.165 & 0.052 & 0.174 & 0.071
              & 0.163 & 0.056 & 0.174 & 0.071 \\
    \qtyrange{5}{10}{\metre\per\second} & 0.250 & 0.078 & 0.191 & 0.079
              & 0.246 & 0.082 & 0.187 & 0.077 \\
    \qty{>10}{\metre\per\second} & 0.227 & 0.083 & 0.220 & 0.100
              & 0.286 & 0.108 & 0.239 & 0.104 \\

    \bottomrule
\end{tabular}

        \tablenoteblock{%
            Each entry reports residual velocity (\unit{\metre\per\second}) / surface mismatch (\unit{\metre}).
            Native references comprise traffic signs and independently verified stationary vehicles from Waymo.  At ego speeds above \qty{10}{\metre\per\second},
            the size-comparable native stationary-vehicle subsets exhibit
            residual velocities of \qty{0.245}{\metre\per\second} (v1.0)
            and \qty{0.274}{\metre\per\second} (XL),
            closely matching the corresponding inserted-object measurements.
        }
    \end{threeparttable}
\end{table}

Future extensions with dynamic object insertions and occupancy annotations could support evaluation of semantic occupancy forecasting~\cite{mohan2026forecastocc}.

\section{Additional Experiments}

\subsection{Cross-Dataset Segmentation Comparison}
\label{ssec:sup-exp-crossdataset}

To contextualize segmentation performance on WayPOP, we compare RbA~\cite{nayal2023rba} across PANIC, COCO with \qty{20}{\percent} withheld classes, and WayPOP-v1.0 in \Cref{tab:cross-dataset-segmentation}.
For PANIC, we use the official Cityscapes-trained RbA checkpoint.
The COCO result follows the corresponding open-set protocol.
RbA achieves unknown PQ of \num{56.5} on PANIC, \num{24.8} on COCO, and \num{31.9} on WayPOP-v1.

\begin{table}[h]
    \centering
    \begin{threeparttable}
        \caption{RbA performance across open-set segmentation benchmarks.}
        \label{tab:cross-dataset-segmentation}

        \small
        \sisetup{
  detect-all,
  table-format=2.1
}
\begin{tabular}{llSS}
    \toprule
    & & \multicolumn{2}{c}{\textbf{PQ} $\uparrow$} \\
    \cmidrule(lr){3-4}
    \textbf{Dataset} & \textbf{Image domain}
    & {Known} & {Unknown} \\
    \midrule
    PANIC & Driving scenes & {--} & 56.5 \\
    COCO (\qty{20}{\percent} withheld classes) & Generic images & 47.0 & 24.8 \\
    WayPOP-v1.0 & Driving scenes & 45.3 & 31.9 \\
    \bottomrule
\end{tabular}

        \tablenoteblock{%
            PANIC reports unknown PQ only.
            Differences in image domains, class taxonomies, and unknown-object
            definitions preclude a controlled comparison of benchmark difficulty.
        }
    \end{threeparttable}
\end{table}

Although PANIC and WayPOP both depict driving scenes, their taxonomies and unknown-object definitions differ, while COCO additionally covers a broader image domain.
These results suggest that WayPOP poses a meaningful challenge even for established open-set segmentation methods.

\subsection{RGB-LiDAR Segmentation}
\label{ssec:sup-exp-lidar}
Additional sensing modalities provide informative cues for scene understanding beyond RGB~\cite{hindel2025label,hurtado2026hyperspectral}. In particular, LiDAR-camera fusion combines geometric and visual information for panoptic segmentation~\cite{mohan2026up}. To explore the utility of the aligned LiDAR measurements provided by WayPOP, we augment RbA~\cite{nayal2023rba} with RGB-LiDAR feature fusion.
Separate ResNet-50~\cite{he2016deep} encoders process RGB images and projected LiDAR range, height, and validity channels.
We combine their features through addition at strides \num{4} and \num{8} and RGB-guided cross-attention at strides \num{16} and \num{32}, retaining the original Mask2Former~\cite{cheng2022masked} decoder and RbA unknown-object scoring.
LiDAR insertion updates range measurements but does not synthesize intensity or elongation for inserted surfaces.
We therefore restrict the LiDAR inputs to range, height, and validity.

\begin{table}[h]
    \centering
    \setlength{\tabcolsep}{20pt} %
    \begin{threeparttable}
        \caption{RGB--LiDAR fusion for open-set panoptic segmentation on WayPOP-v1.}
        \label{tab:rgb-lidar-segmentation}
        \small
        \sisetup{
  detect-all,
  table-format=2.1
}
\begin{tabular}{ccSS}
    \toprule
    \multicolumn{2}{c}{\textbf{Input}} & \multicolumn{2}{c}{\textbf{PQ} $\uparrow$} \\
    \cmidrule(lr){1-2}\cmidrule(lr){3-4}
    RGB & LiDAR & {Known} & {Unknown} \\
    \midrule
    \cmark & \xmark & 45.3 & 31.9 \\
    \cmark & \cmark & \bfseries 46.8 & \bfseries 34.1 \\
    \bottomrule
\end{tabular}

        \tablenoteblock{%
            Both configurations use RbA for unknown-object scoring.
        }
    \end{threeparttable}
\end{table}

As shown in \Cref{tab:rgb-lidar-segmentation}, adding LiDAR improves known PQ from \num{45.3} to \num{46.8} and unknown PQ from \num{31.9} to \num{34.1}.
These exploratory results suggest that geometric information can complement camera inputs for both known and unknown segmentation.

\subsection{Additional Frontal-View Tracking Results}
\label{ssec:sup-exp-fv-ospt}

We compare frontal-view tracking performance for the strongest open-set segmentation methods across WayPOP-v1.0 and WayPOP-XL in \Cref{tab:results-track-fron-supp}.
This comparison examines how the two asset sources affect segmentation and association performance.

\begin{table}[h]
    \centering
    \begin{threeparttable}
        \caption{FV-OSPT results on open-set methods across WayPOP-v1 and WayPOP-XL.}
        \label{tab:results-track-fron-supp}

        \small
        \centering
\footnotesize
\setlength{\tabcolsep}{3pt}  %

\sisetup{
  detect-all,
  table-number-alignment=center,
  round-mode=places,
  round-precision=1
}

\begin{tabular}{
l l
*{12}{S[table-format=2.1]}
}
\toprule
& &
\multicolumn{6}{c}{\textbf{WayPOP-v1}} &
\multicolumn{6}{c}{\textbf{WayPOP-XL}} \\
\cmidrule(lr){3-8}
\cmidrule(lr){9-14}

& &
\multicolumn{3}{c}{Known} &
\multicolumn{3}{c}{Unknown} &
\multicolumn{3}{c}{Known} &
\multicolumn{3}{c}{Unknown} \\
\cmidrule(lr){3-5}
\cmidrule(lr){6-8}
\cmidrule(lr){9-11}
\cmidrule(lr){12-14}

& \textbf{Tracker}
& {STQ $\uparrow$} & {AQ $\uparrow$} & {mIoU $\uparrow$}
& {STQ $\uparrow$} & {AQ $\uparrow$} & {mIoU $\uparrow$}
& {STQ $\uparrow$} & {AQ $\uparrow$} & {mIoU $\uparrow$}
& {STQ $\uparrow$} & {AQ $\uparrow$} & {mIoU $\uparrow$} \\

\midrule

\multirow{4}{*}{\rotatebox[origin=c]{90}{\textbf{PoDS}~\cite{mohan2024panoptic}}}
& TrackTrack~\cite{shim2025focusing}
    & 16.7 & 5.6 & 49.6
    & 26.4 & 16.0 & 43.5
    & 16.6 & 5.3 & 51.7
    & 31.0 & 17.0 & 56.5 \\

& BoT-SORT~\cite{aharon2022bot}
    & 16.5 & 5.5 & 49.6
    & 25.4 & 14.8 & 43.5
    & 16.7 & 5.4 & 51.7
    & 29.7 & 15.6 & 56.5 \\

& MASA~\cite{li2024matching}
    & 15.8 & 5.0 & 49.6
    & \bfseries 30.0 & \bfseries 20.7 & 43.5
    & 15.3 & 4.5 & 51.7
    & \bfseries 33.8 & \bfseries 20.3 & 56.5 \\

& DEVA~\cite{cheng2023tracking}
    & \bfseries 19.1 & \bfseries 7.4 & 49.6
    & 29.0 & 19.3 & 43.5
    & \bfseries 18.5 & \bfseries 6.6 & 51.7
    & 32.2 & 18.4 & 56.5 \\

\arrayrulecolor{gray}\midrule\arrayrulecolor{black}

\multirow{4}{*}{\rotatebox[origin=c]{90}{\textbf{RbA}~\cite{nayal2023rba}}}
& TrackTrack~\cite{shim2025focusing}
    & 18.4 & 6.8 & 49.4
    & 26.3 & 18.1 & 38.3
    & 18.1 & 6.4 & 51.5
    & 32.1 & 18.9 & 54.4 \\

& BoT-SORT~\cite{aharon2022bot}
    & 17.4 & 6.1 & 49.4
    & 25.2 & 16.6 & 38.3
    & 17.3 & 5.8 & 51.5
    & 30.6 & 17.2 & 54.4 \\

& MASA~\cite{li2024matching}
    & 16.2 & 5.3 & 49.4
    & \bfseries 29.6 & \bfseries 22.9 & 38.3
    & 16.1 & 5.0 & 51.5
    & \bfseries 34.7 & \bfseries 22.1 & 54.4 \\

& DEVA~\cite{cheng2023tracking}
    & \bfseries 20.6 & \bfseries 8.5 & 49.4
    & 28.7 & 21.5 & 38.3
    & \bfseries 19.8 & \bfseries 7.6 & 51.5
    & 33.2 & 20.3 & 54.4 \\

\bottomrule
\end{tabular}

        \tablenoteblock{%
        }
    \end{threeparttable}
\end{table}

Known-class performance remains similar across the two variants, and MASA achieves the highest unknown STQ and AQ for both open-set models on both splits.
PoDS achieves higher unknown STQ than RbA on WayPOP-v1.0 for every tracker, whereas RbA leads on WayPOP-XL.
Unknown STQ is consistently higher on WayPOP-XL, accompanied by substantially higher unknown mIoU, while AQ changes are smaller and vary across trackers.
For example, RbA with MASA achieves unknown STQ of \num{29.6} on v1.0 and \num{34.7} on XL, with AQ of \num{22.9} and \num{22.1}, respectively.
The higher unknown tracking scores on XL therefore primarily reflect improved segmentation rather than more accurate association.

\subsection{Perfect-Mask Tracking Analysis}
\label{ssec:sup-exp-perfect-mask}

To isolate association difficulty from segmentation errors, we evaluate frontal-view tracking with ground-truth masks and classes.
Even with perfect frame-local predictions, identities must be maintained under viewpoint changes, occlusion, and image-space displacement induced by ego motion.
For adjacent observations of the same unknown object, the median mask IoU is \num{0.1539} and the median centroid displacement is \num{13.84} pixels.

We additionally evaluate a privileged static-object prior that uses the known staticity of inserted objects to relink unknown-object tracklets.
Ground-contact points are estimated from the bottom
\qty{2}{\percent} of each mask, backprojected onto the ground
plane using camera calibration, and transformed into a common
world coordinate frame using the recorded ego poses.
Non-overlapping tracklets are relinked using Hungarian assignment,
with a maximum temporal gap of \qty{1}{\second},
world-space separation of \qty{1.5}{\metre},
reprojection error of \num{120} pixels per direction,
and symmetric mask-area ratio of \num{4}.
Ambiguous assignments are rejected using a minimum cost margin
of \num{0.15}.

\begin{table}[h]
    \centering
    \begin{threeparttable}
        \caption{Frontal-view tracking with ground-truth masks and classes on WayPOP-v1.}
        \label{tab:perfect-mask-tracking}
        \small
        \sisetup{
  detect-all,
  table-format=2.2
}
\begin{tabular}{lc*{4}{S}}
    \toprule
    & & \multicolumn{2}{c}{\textbf{Known}}
      & \multicolumn{2}{c}{\textbf{Unknown}} \\
    \cmidrule(lr){3-4}\cmidrule(lr){5-6}
    \textbf{Tracker} & \textbf{Static prior}
        & {STQ $\uparrow$} & {AQ $\uparrow$}
        & {STQ $\uparrow$} & {AQ $\uparrow$} \\
    \midrule

    DEVA~\cite{cheng2023tracking} & \xmark
        & 88.64 & 78.57 & 77.37 & 59.86 \\
         & \cmark
        & 88.64 & 78.57 & 82.68 & 68.35 \\

    \arrayrulecolor{gray}
    \midrule
    \arrayrulecolor{black}

    SAM 3.1~\cite{carion2026sam3} & \xmark
        & 88.12 & 77.66 & 56.89 & 32.36 \\
            & \cmark
        & 88.12 & 77.66 & 77.43 & 59.95 \\

    \bottomrule
\end{tabular}

        \tablenoteblock{%
            All configurations use perfect frame-local masks and classes
            (mIoU = \num{100}).
            The static prior uses ego-motion compensation to relink
            unknown-object tracklets; known-class results remain unchanged
            by construction.
        }
    \end{threeparttable}
\end{table}

As shown in \Cref{tab:perfect-mask-tracking}, unknown AQ improves from \num{59.86} to \num{68.35} for DEVA and from \num{32.36} to \num{59.95} for SAM 3.1 when using this prior.
These gains indicate that static-scene geometry can help recover association continuity, while the remaining errors show that perfect masks and known staticity do not fully resolve tracking.

\clearpage
\pdfbookmark[1]{References}{references}
{
\small
\bibliography{main}

\begin{thebibliography}{72}
\providecommand{\natexlab}[1]{#1}
\providecommand{\url}[1]{\texttt{#1}}
\expandafter\ifx\csname urlstyle\endcsname\relax
  \providecommand{\doi}[1]{doi: #1}\else
  \providecommand{\doi}{doi: \begingroup \urlstyle{rm}\Url}\fi

\bibitem[Luz et~al.(2026)Luz, Mohan, N{\"u}rnberg, Miron, Cattaneo, and
  Valada]{luz2026latent}
Maximilian Luz, Rohit Mohan, Thomas N{\"u}rnberg, Yakov Miron, Daniele
  Cattaneo, and Abhinav Valada.
\newblock Latent gaussian splatting for 4d panoptic occupancy tracking.
\newblock \emph{IEEE Robotics and Automation Letters}, 2026.

\bibitem[Fong et~al.(2022)Fong, Mohan, Hurtado, Zhou, Caesar, Beijbom, and
  Valada]{fong2022panoptic}
Whye~Kit Fong, Rohit Mohan, Juana~Valeria Hurtado, Lubing Zhou, Holger Caesar,
  Oscar Beijbom, and Abhinav Valada.
\newblock Panoptic {nuScenes}: A large-scale benchmark for {LiDAR} panoptic
  segmentation and tracking.
\newblock \emph{IEEE Robotics and Automation Letters (RA-L)}, 7\penalty0
  (2):\penalty0 3795--3802, 2022.

\bibitem[Hurtado et~al.(2020)Hurtado, Mohan, Burgard, and
  Valada]{hurtado2020mopt}
Juana~Valeria Hurtado, Rohit Mohan, Wolfram Burgard, and Abhinav Valada.
\newblock {MOPT}: Multi-object panoptic tracking.
\newblock In \emph{IEEE/CVF Conference on Computer Vision and Pattern
  Recognition Workshops (CVPRW)}, 2020.

\bibitem[Kim et~al.(2020)Kim, Woo, Lee, and Kweon]{kim2020video}
Dahun Kim, Sanghyun Woo, Joon-Young Lee, and In~So Kweon.
\newblock Video panoptic segmentation.
\newblock In \emph{IEEE/CVF Conference on Computer Vision and Pattern
  Recognition (CVPR)}, 2020.

\bibitem[Mei et~al.(2022)Mei, Zhu, Yan, Yan, Qiao, Chen, and
  Kretzschmar]{mei2022pvps}
Jieru Mei, Alex~Zihao Zhu, Xinchen Yan, Hang Yan, Siyuan Qiao, Liang-Chieh
  Chen, and Henrik Kretzschmar.
\newblock {Waymo Open Dataset}: Panoramic video panoptic segmentation.
\newblock In \emph{European Conference on Computer Vision (ECCV)}, 2022.

\bibitem[Pinggera et~al.(2016)Pinggera, Ramos, Gehrig, Franke, Rother, and
  Mester]{pinggera2016lost}
Peter Pinggera, Sebastian Ramos, Stefan Gehrig, Uwe Franke, Carsten Rother, and
  Rudolf Mester.
\newblock {Lost and Found}: Detecting small road hazards for self-driving
  vehicles.
\newblock In \emph{IEEE/RSJ International Conference on Intelligent Robots and
  Systems (IROS)}, 2016.

\bibitem[Chan et~al.(2021)Chan, Lis, Uhlemeyer, Blum, Honari, Siegwart, Fua,
  Salzmann, and Rottmann]{chan2021segmentmeifyoucan}
Robin Chan, Krzysztof Lis, Svenja Uhlemeyer, Hermann Blum, Sina Honari, Roland
  Siegwart, Pascal Fua, Mathieu Salzmann, and Matthias Rottmann.
\newblock {SegmentMeIfYouCan}: A benchmark for anomaly segmentation.
\newblock In \emph{Neural Information Processing Systems Track on Datasets and
  Benchmarks (NeurIPS D\&B)}, 2021.

\bibitem[Hwang et~al.(2021)Hwang, Oh, Lee, and Han]{hwang2021exemplar}
Jaedong Hwang, Seoung~Wug Oh, Joon-Young Lee, and Bohyung Han.
\newblock Exemplar-based open-set panoptic segmentation network.
\newblock In \emph{IEEE/CVF Conference on Computer Vision and Pattern
  Recognition (CVPR)}, 2021.

\bibitem[Sodano et~al.(2024)Sodano, Magistri, Behley, and
  Stachniss]{sodano2024open}
Matteo Sodano, Federico Magistri, Jens Behley, and Cyrill Stachniss.
\newblock Open-world panoptic segmentation.
\newblock \emph{arXiv preprint arXiv:2412.12740}, 2024.
\newblock Accepted to the International Journal of Robotics Research (IJRR).

\bibitem[Mohan et~al.(2024{\natexlab{a}})Mohan, Kumaraswamy, Hurtado, Petek,
  and Valada]{mohan2024panoptic}
Rohit Mohan, Kiran Kumaraswamy, Juana~Valeria Hurtado, K{\"u}rsat Petek, and
  Abhinav Valada.
\newblock Panoptic out-of-distribution segmentation.
\newblock \emph{IEEE Robotics and Automation Letters (RA-L)}, 9\penalty0
  (5):\penalty0 4075--4082, 2024{\natexlab{a}}.

\bibitem[Bogdoll et~al.(2024)Bogdoll, Hamdard, R{\"o}{\ss}ler, Geisler, Bayram,
  Wang, Imhof, de~Campos, Tabarov, Yang, Gontscharow, Gottschalk, and
  Z{\"o}llner]{bogdoll2024anovox}
Daniel Bogdoll, Iramm Hamdard, Lukas~Namgyu R{\"o}{\ss}ler, Felix Geisler,
  Muhammed Bayram, Felix Wang, Jan Imhof, Miguel de~Campos, Anushervon Tabarov,
  Yitian Yang, Martin Gontscharow, Hanno Gottschalk, and J.~Marius Z{\"o}llner.
\newblock {AnoVox}: A benchmark for multimodal anomaly detection in autonomous
  driving.
\newblock In \emph{European Conference on Computer Vision Workshops (ECCVW)},
  2024.

\bibitem[Nekrasov et~al.(2025)Nekrasov, Burdorf, Worrall, Leibe, and
  Perez]{nekrasov2025spotting}
Alexey Nekrasov, Malcolm Burdorf, Stewart Worrall, Bastian Leibe, and Julie
  Stephany~Berrio Perez.
\newblock Spotting the unexpected ({STU}): A {3D} {LiDAR} dataset for anomaly
  segmentation in autonomous driving.
\newblock In \emph{IEEE/CVF Conference on Computer Vision and Pattern
  Recognition (CVPR)}, 2025.

\bibitem[Deitke et~al.(2023{\natexlab{a}})Deitke, Schwenk, Salvador, Weihs,
  Michel, VanderBilt, Schmidt, Ehsani, Kembhavi, and
  Farhadi]{deitke2023objaverse}
Matt Deitke, Dustin Schwenk, Jordi Salvador, Luca Weihs, Oscar Michel, Eli
  VanderBilt, Ludwig Schmidt, Kiana Ehsani, Aniruddha Kembhavi, and Ali
  Farhadi.
\newblock {Objaverse}: A universe of annotated {3D} objects.
\newblock In \emph{IEEE/CVF Conference on Computer Vision and Pattern
  Recognition (CVPR)}, 2023{\natexlab{a}}.

\bibitem[Mohan et~al.(2024{\natexlab{b}})Mohan, Arce, Mokhtar, Cattaneo, and
  Valada]{mohan2024syn}
Rohit Mohan, Jose Arce, Sassan Mokhtar, Daniele Cattaneo, and Abhinav Valada.
\newblock Syn-mediverse: a multimodal synthetic dataset for intelligent scene
  understanding of healthcare facilities.
\newblock \emph{IEEE Robotics and Automation Letters}, 9\penalty0 (8):\penalty0
  7094--7101, 2024{\natexlab{b}}.

\bibitem[Sekkat et~al.(2024)Sekkat, Mohan, Sawade, Matthes, and
  Valada]{sekkat2024amodalsynthdrive}
Ahmed~Rida Sekkat, Rohit Mohan, Oliver Sawade, Elmar Matthes, and Abhinav
  Valada.
\newblock Amodalsynthdrive: A synthetic amodal perception dataset for
  autonomous driving.
\newblock \emph{IEEE Robotics and Automation Letters}, 9\penalty0
  (11):\penalty0 9597--9604, 2024.

\bibitem[Blum et~al.(2019)Blum, Sarlin, Nieto, Siegwart, and
  Cadena]{blum2019fishyscapes}
Hermann Blum, Paul-Edouard Sarlin, Juan Nieto, Roland Siegwart, and Cesar
  Cadena.
\newblock {Fishyscapes}: A benchmark for safe semantic segmentation in
  autonomous driving.
\newblock In \emph{IEEE/CVF International Conference on Computer Vision
  Workshops (ICCVW)}, 2019.

\bibitem[Cordts et~al.(2016)Cordts, Omran, Ramos, Rehfeld, Enzweiler, Benenson,
  Franke, Roth, and Schiele]{cordts2016cityscapes}
Marius Cordts, Mohamed Omran, Sebastian Ramos, Timo Rehfeld, Markus Enzweiler,
  Rodrigo Benenson, Uwe Franke, Stefan Roth, and Bernt Schiele.
\newblock The {Cityscapes} dataset for semantic urban scene understanding.
\newblock In \emph{IEEE Conference on Computer Vision and Pattern Recognition
  (CVPR)}, 2016.

\bibitem[Hendrycks et~al.(2022)Hendrycks, Basart, Mazeika, Zou, Kwon,
  Mostajabi, Steinhardt, and Song]{basart2022scaling}
Dan Hendrycks, Steven Basart, Mantas Mazeika, Andy Zou, Joseph Kwon,
  Mohammadreza Mostajabi, Jacob Steinhardt, and Dawn Song.
\newblock Scaling out-of-distribution detection for real-world settings.
\newblock In \emph{International Conference on Machine Learning (ICML)}, 2022.

\bibitem[Yu et~al.(2020)Yu, Chen, Wang, Xian, Chen, Liu, Madhavan, and
  Darrell]{yu2020bdd100k}
Fisher Yu, Haofeng Chen, Xin Wang, Wenqi Xian, Yingying Chen, Fangchen Liu,
  Vashisht Madhavan, and Trevor Darrell.
\newblock {BDD100K}: A diverse driving dataset for heterogeneous multitask
  learning.
\newblock In \emph{IEEE/CVF Conference on Computer Vision and Pattern
  Recognition (CVPR)}, 2020.

\bibitem[Maag et~al.(2022)Maag, Chan, Uhlemeyer, Kowol, and
  Gottschalk]{maag2022two}
Kira Maag, Robin Chan, Svenja Uhlemeyer, Kamil Kowol, and Hanno Gottschalk.
\newblock Two video data sets for tracking and retrieval of out of distribution
  objects.
\newblock In \emph{Asian Conference on Computer Vision (ACCV)}, 2022.

\bibitem[Singh et~al.(2020)Singh, Kamireddypalli, Gandhi, and
  Krishna]{singh2020lidar}
Aasheesh Singh, Aditya Kamireddypalli, Vineet Gandhi, and K~Madhava Krishna.
\newblock {LiDAR} guided small obstacle segmentation.
\newblock In \emph{IEEE/RSJ International Conference on Intelligent Robots and
  Systems (IROS)}, 2020.

\bibitem[Bevandi{\'c} et~al.(2019)Bevandi{\'c}, Kre{\v{s}}o, Or{\v{s}}i{\'c},
  and {\v{S}}egvi{\'c}]{bevandic2019simultaneous}
Petra Bevandi{\'c}, Ivan Kre{\v{s}}o, Marin Or{\v{s}}i{\'c}, and Sini{\v{s}}a
  {\v{S}}egvi{\'c}.
\newblock Simultaneous semantic segmentation and outlier detection in presence
  of domain shift.
\newblock In \emph{German Conference on Pattern Recognition (GCPR)}, 2019.

\bibitem[Grci{\'c} et~al.(2021)Grci{\'c}, Bevandi{\'c}, and
  {\v{S}}egvi{\'c}]{grcic2020dense}
Matej Grci{\'c}, Petra Bevandi{\'c}, and Sini{\v{s}}a {\v{S}}egvi{\'c}.
\newblock Dense open-set recognition with synthetic outliers generated by {Real
  NVP}.
\newblock In \emph{International Conference on Computer Vision Theory and
  Applications (VISAPP)}, 2021.

\bibitem[Neuhold et~al.(2017)Neuhold, Ollmann, Rota~Bulo, and
  Kontschieder]{neuhold2017mapillary}
Gerhard Neuhold, Tobias Ollmann, Samuel Rota~Bulo, and Peter Kontschieder.
\newblock The {Mapillary Vistas} dataset for semantic understanding of street
  scenes.
\newblock In \emph{IEEE International Conference on Computer Vision (ICCV)},
  2017.

\bibitem[Dosovitskiy et~al.(2017)Dosovitskiy, Ros, Codevilla, Lopez, and
  Koltun]{dosovitskiy2017carla}
Alexey Dosovitskiy, German Ros, Felipe Codevilla, Antonio Lopez, and Vladlen
  Koltun.
\newblock {CARLA}: An open urban driving simulator.
\newblock In \emph{Conference on Robot Learning (CoRL)}, 2017.

\bibitem[Kirillov et~al.(2019)Kirillov, He, Girshick, Rother, and
  Doll{\'a}r]{kirillov2019panoptic}
Alexander Kirillov, Kaiming He, Ross Girshick, Carsten Rother, and Piotr
  Doll{\'a}r.
\newblock Panoptic segmentation.
\newblock In \emph{IEEE/CVF Conference on Computer Vision and Pattern
  Recognition (CVPR)}, 2019.

\bibitem[Lin et~al.(2014)Lin, Maire, Belongie, Hays, Perona, Ramanan,
  Doll{\'a}r, and Zitnick]{lin2014microsoft}
Tsung-Yi Lin, Michael Maire, Serge Belongie, James Hays, Pietro Perona, Deva
  Ramanan, Piotr Doll{\'a}r, and C.~Lawrence Zitnick.
\newblock Microsoft {COCO}: Common objects in context.
\newblock In \emph{European Conference on Computer Vision (ECCV)}, 2014.

\bibitem[Mohan and Valada(2021)]{mohan2021efficientps}
Rohit Mohan and Abhinav Valada.
\newblock {EfficientPS}: Efficient panoptic segmentation.
\newblock \emph{International Journal of Computer Vision (IJCV)}, 129\penalty0
  (5):\penalty0 1551--1579, 2021.

\bibitem[Cheng et~al.(2020)Cheng, Collins, Zhu, Liu, Huang, Adam, and
  Chen]{cheng2020panoptic}
Bowen Cheng, Maxwell~D Collins, Yukun Zhu, Ting Liu, Thomas~S Huang, Hartwig
  Adam, and Liang-Chieh Chen.
\newblock {Panoptic-DeepLab}: A simple, strong, and fast baseline for bottom-up
  panoptic segmentation.
\newblock In \emph{IEEE/CVF Conference on Computer Vision and Pattern
  Recognition (CVPR)}, 2020.

\bibitem[Cheng et~al.(2022)Cheng, Misra, Schwing, Kirillov, and
  Girdhar]{cheng2022masked}
Bowen Cheng, Ishan Misra, Alexander~G Schwing, Alexander Kirillov, and Rohit
  Girdhar.
\newblock Masked-attention mask transformer for universal image segmentation.
\newblock In \emph{IEEE/CVF Conference on Computer Vision and Pattern
  Recognition (CVPR)}, 2022.

\bibitem[Weber et~al.(2021)Weber, Xie, Collins, Zhu, Voigtlaender, Adam, Green,
  Geiger, Leibe, Cremers, O{\v{s}}ep, Leal-Taix{\'e}, and Chen]{weber2021step}
Mark Weber, Jun Xie, Maxwell Collins, Yukun Zhu, Paul Voigtlaender, Hartwig
  Adam, Bradley Green, Andreas Geiger, Bastian Leibe, Daniel Cremers,
  Aljo{\v{s}}a O{\v{s}}ep, Laura Leal-Taix{\'e}, and Liang-Chieh Chen.
\newblock {STEP}: Segmenting and tracking every pixel.
\newblock In \emph{Neural Information Processing Systems Track on Datasets and
  Benchmarks (NeurIPS D\&B)}, 2021.

\bibitem[Jaus et~al.(2023)Jaus, Yang, and Stiefelhagen]{jaus2023panoramic}
Alexander Jaus, Kailun Yang, and Rainer Stiefelhagen.
\newblock Panoramic panoptic segmentation: Insights into surrounding parsing
  for mobile agents via unsupervised contrastive learning.
\newblock \emph{IEEE Transactions on Intelligent Transportation Systems
  (T-ITS)}, 24\penalty0 (4):\penalty0 4438--4453, 2023.

\bibitem[Mohan et~al.(2025)Mohan, Hindel, Drews, Gl{\"a}ser, Cattaneo, and
  Valada]{mohan2025open}
Rohit Mohan, Julia Hindel, Florian Drews, Claudius Gl{\"a}ser, Daniele
  Cattaneo, and Abhinav Valada.
\newblock Open-set lidar panoptic segmentation guided by uncertainty-aware
  learning.
\newblock In \emph{2025 IEEE/RSJ International Conference on Intelligent Robots
  and Systems (IROS)}, pages 2224--2231. IEEE, 2025.

\bibitem[Le et~al.(2024)Le, Gou, Datta, Shi, Reid, Cai, and
  Rezatofighi]{le2024jrdb}
Duy~Tho Le, Chenhui Gou, Stavya Datta, Hengcan Shi, Ian Reid, Jianfei Cai, and
  Hamid Rezatofighi.
\newblock {JRDB-PanoTrack}: An open-world panoptic segmentation and tracking
  robotic dataset in crowded human environments.
\newblock In \emph{IEEE/CVF Conference on Computer Vision and Pattern
  Recognition (CVPR)}, 2024.

\bibitem[Li et~al.(2022)Li, Chen, Wang, Hong, Ye, Han, Chen, Zhang, Xu, Yeung,
  Liang, Li, and Xu]{li2022coda}
Kaican Li, Kai Chen, Haoyu Wang, Lanqing Hong, Chaoqiang Ye, Jianhua Han,
  Yukuai Chen, Wei Zhang, Chunjing Xu, Dit-Yan Yeung, Xiaodan Liang, Zhenguo
  Li, and Hang Xu.
\newblock {CODA}: A real-world road corner case dataset for object detection in
  autonomous driving.
\newblock In \emph{European Conference on Computer Vision (ECCV)}, 2022.

\bibitem[Xia et~al.(2025)Xia, Li, Lin, Wang, Wang, and Yang]{xia2025openad}
Zhongyu Xia, Jishuo Li, Zhiwei Lin, Xinhao Wang, Yongtao Wang, and Ming-Hsuan
  Yang.
\newblock {OpenAD}: Open-world autonomous driving benchmark for {3D} object
  detection.
\newblock In \emph{Advances in Neural Information Processing Systems, Datasets
  and Benchmarks Track (NeurIPS D\&B)}, 2025.

\bibitem[Gasperini et~al.(2023)Gasperini, Marcos-Ramiro, Schmidt, Navab, Busam,
  and Tombari]{gasperini2023segmenting}
Stefano Gasperini, Alvaro Marcos-Ramiro, Michael Schmidt, Nassir Navab,
  Benjamin Busam, and Federico Tombari.
\newblock Segmenting known objects and unseen unknowns without prior knowledge.
\newblock In \emph{IEEE/CVF International Conference on Computer Vision
  (ICCV)}, 2023.

\bibitem[Schmidt et~al.(2025)Schmidt, K{\"o}rner, Fuchsgruber, Gasperini,
  Tombari, and G{\"u}nnemann]{schmidt2025prior2former}
Sebastian Schmidt, Julius K{\"o}rner, Dominik Fuchsgruber, Stefano Gasperini,
  Federico Tombari, and Stephan G{\"u}nnemann.
\newblock {Prior2Former} -- evidential modeling of mask transformers for
  assumption-free open-world panoptic segmentation.
\newblock In \emph{IEEE/CVF International Conference on Computer Vision
  (ICCV)}, 2025.

\bibitem[Rai et~al.(2024)Rai, Cermelli, Caputo, and
  Masone]{rai2024mask2anomaly}
Shyam~Nandan Rai, Fabio Cermelli, Barbara Caputo, and Carlo Masone.
\newblock {Mask2Anomaly}: Mask transformer for universal open-set segmentation.
\newblock \emph{IEEE Transactions on Pattern Analysis and Machine Intelligence
  (T-PAMI)}, 46\penalty0 (12):\penalty0 9286--9302, 2024.

\bibitem[Nayal et~al.(2023)Nayal, Yavuz, Henriques, and
  G{\"u}ney]{nayal2023rba}
Nazir Nayal, M{\i}sra Yavuz, Jo{\~a}o~F. Henriques, and Fatma G{\"u}ney.
\newblock {RbA}: Segmenting unknown regions rejected by all.
\newblock In \emph{IEEE/CVF International Conference on Computer Vision
  (ICCV)}, 2023.

\bibitem[Lu et~al.(2026)Lu, Mohan, Drews, Miron, and Valada]{lu2026hyp2former}
Yao Lu, Rohit Mohan, Florian Drews, Yakov Miron, and Abhinav Valada.
\newblock {Hyp2Former}: Hierarchy-aware hyperbolic embeddings for open-set
  panoptic segmentation.
\newblock \emph{arXiv preprint arXiv:2605.02580}, 2026.

\bibitem[Yu et~al.(2023)Yu, He, Deng, Shen, and Chen]{yu2023fcclip}
Qihang Yu, Ju~He, Xueqing Deng, Xiaohui Shen, and Liang-Chieh Chen.
\newblock Convolutions die hard: Open-vocabulary segmentation with single
  frozen convolutional {CLIP}.
\newblock In \emph{Advances in Neural Information Processing Systems
  (NeurIPS)}, 2023.

\bibitem[Jiao et~al.(2024)Jiao, Zhu, Huang, Zhao, Wei, and
  Shi]{jiao2024maftplus}
Siyu Jiao, Hongguang Zhu, Jiannan Huang, Yao Zhao, Yunchao Wei, and Humphrey
  Shi.
\newblock Collaborative vision-text representation optimizing for
  open-vocabulary segmentation.
\newblock In \emph{European Conference on Computer Vision (ECCV)}, 2024.

\bibitem[Dong et~al.(2023)Dong, Bao, Zheng, Zhang, Chen, Yang, Zeng, Zhang,
  Yuan, Chen, Wen, and Yu]{dong2023maskclip}
Xiaoyi Dong, Jianmin Bao, Yinglin Zheng, Ting Zhang, Dongdong Chen, Hao Yang,
  Ming Zeng, Weiming Zhang, Lu~Yuan, Dong Chen, Fang Wen, and Nenghai Yu.
\newblock {MaskCLIP}: Masked self-distillation advances contrastive
  language-image pretraining.
\newblock In \emph{IEEE/CVF Conference on Computer Vision and Pattern
  Recognition (CVPR)}, 2023.

\bibitem[Xu et~al.(2023)Xu, Liu, Vahdat, Byeon, Wang, and
  De~Mello]{xu2023odise}
Jiarui Xu, Sifei Liu, Arash Vahdat, Wonmin Byeon, Xiaolong Wang, and Shalini
  De~Mello.
\newblock Open-vocabulary panoptic segmentation with text-to-image diffusion
  models.
\newblock In \emph{IEEE/CVF Conference on Computer Vision and Pattern
  Recognition (CVPR)}, 2023.

\bibitem[Shim et~al.(2025)Shim, Ko, Yang, and Kim]{shim2025focusing}
Kyujin Shim, Kangwook Ko, Yujin Yang, and Changick Kim.
\newblock Focusing on tracks for online multi-object tracking.
\newblock In \emph{IEEE/CVF Conference on Computer Vision and Pattern
  Recognition (CVPR)}, 2025.

\bibitem[Aharon et~al.(2022)Aharon, Orfaig, and Bobrovsky]{aharon2022bot}
Nir Aharon, Roy Orfaig, and Ben-Zion Bobrovsky.
\newblock {BoT-SORT}: Robust associations multi-pedestrian tracking.
\newblock \emph{arXiv preprint arXiv:2206.14651}, 2022.

\bibitem[Zhang et~al.(2022)Zhang, Sun, Jiang, Yu, Weng, Yuan, Luo, Liu, and
  Wang]{zhang2022bytetrack}
Yifu Zhang, Peize Sun, Yi~Jiang, Dongdong Yu, Fucheng Weng, Zehuan Yuan, Ping
  Luo, Wenyu Liu, and Xinggang Wang.
\newblock {ByteTrack}: Multi-object tracking by associating every detection
  box.
\newblock In \emph{European Conference on Computer Vision (ECCV)}, 2022.

\bibitem[Cao et~al.(2023)Cao, Pang, Weng, Khirodkar, and
  Kitani]{cao2023observation}
Jinkun Cao, Jiangmiao Pang, Xinshuo Weng, Rawal Khirodkar, and Kris Kitani.
\newblock Observation-centric {SORT}: Rethinking {SORT} for robust multi-object
  tracking.
\newblock In \emph{IEEE/CVF Conference on Computer Vision and Pattern
  Recognition (CVPR)}, 2023.

\bibitem[Li et~al.(2024)Li, Ke, Danelljan, Piccinelli, Segu, Van~Gool, and
  Yu]{li2024matching}
Siyuan Li, Lei Ke, Martin Danelljan, Luigi Piccinelli, Mattia Segu, Luc
  Van~Gool, and Fisher Yu.
\newblock Matching anything by segmenting anything.
\newblock In \emph{IEEE/CVF Conference on Computer Vision and Pattern
  Recognition (CVPR)}, 2024.

\bibitem[Cheng et~al.(2023)Cheng, Oh, Price, Schwing, and
  Lee]{cheng2023tracking}
Ho~Kei Cheng, Seoung~Wug Oh, Brian Price, Alexander Schwing, and Joon-Young
  Lee.
\newblock Tracking anything with decoupled video segmentation.
\newblock In \emph{IEEE/CVF International Conference on Computer Vision
  (ICCV)}, 2023.

\bibitem[Jiang et~al.(2026)Jiang, Wang, Zhao, Li, and Jiang]{jiang2026sam2mot}
Junjie Jiang, Zelin Wang, Manqi Zhao, Yin Li, and Dongsheng Jiang.
\newblock {SAM2MOT}: A novel paradigm of multi-object tracking by segmentation.
\newblock In \emph{AAAI Conference on Artificial Intelligence (AAAI)}, 2026.

\bibitem[Luz et~al.(2024)Luz, Mohan, Sekkat, Sawade, Matthes, Brox, and
  Valada]{luz2024amodal}
Maximilian Luz, Rohit Mohan, Ahmed~Rida Sekkat, Oliver Sawade, Elmar Matthes,
  Thomas Brox, and Abhinav Valada.
\newblock Amodal optical flow.
\newblock In \emph{2024 IEEE International Conference on Robotics and
  Automation (ICRA)}, pages 14677--14684. IEEE, 2024.

\bibitem[Geiger et~al.(2012)Geiger, Lenz, and Urtasun]{geiger2012kitti}
Andreas Geiger, Philip Lenz, and Raquel Urtasun.
\newblock Are we ready for autonomous driving? {The} {KITTI} vision benchmark
  suite.
\newblock In \emph{IEEE Conference on Computer Vision and Pattern Recognition
  (CVPR)}, 2012.

\bibitem[Sun et~al.(2020)Sun, Kretzschmar, Dotiwalla, Chouard, Patnaik, Tsui,
  Guo, Zhou, Chai, Caine, Vasudevan, Han, Ngiam, Zhao, Timofeev, Ettinger,
  Krivokon, Gao, Joshi, Zhang, Shlens, Chen, and Anguelov]{sun2020waymo}
Pei Sun, Henrik Kretzschmar, Xerxes Dotiwalla, Aur{\'e}lien Chouard, Vijaysai
  Patnaik, Paul Tsui, James Guo, Yin Zhou, Yuning Chai, Benjamin Caine, Vijay
  Vasudevan, Wei Han, Jiquan Ngiam, Hang Zhao, Aleksei Timofeev, Scott
  Ettinger, Maxim Krivokon, Amy Gao, Aditya Joshi, Yu~Zhang, Jonathon Shlens,
  Zhifeng Chen, and Dragomir Anguelov.
\newblock Scalability in perception for autonomous driving: {Waymo Open
  Dataset}.
\newblock In \emph{IEEE/CVF Conference on Computer Vision and Pattern
  Recognition (CVPR)}, 2020.

\bibitem[Deitke et~al.(2023{\natexlab{b}})Deitke, Liu, Wallingford, Ngo,
  Michel, Kusupati, Fan, Laforte, Voleti, Gadre, VanderBilt, Kembhavi,
  Vondrick, Gkioxari, Ehsani, Schmidt, and Farhadi]{deitke2023objaverseXL}
Matt Deitke, Ruoshi Liu, Matthew Wallingford, Huong Ngo, Oscar Michel, Aditya
  Kusupati, Alan Fan, Christian Laforte, Vikram Voleti, Samir~Yitzhak Gadre,
  Eli VanderBilt, Aniruddha Kembhavi, Carl Vondrick, Georgia Gkioxari, Kiana
  Ehsani, Ludwig Schmidt, and Ali Farhadi.
\newblock {Objaverse-XL}: A universe of {10M+} {3D} objects.
\newblock In \emph{Advances in Neural Information Processing Systems, Datasets
  and Benchmarks Track (NeurIPS D\&B)}, 2023{\natexlab{b}}.

\bibitem[Gupta et~al.(2019)Gupta, Doll{\'a}r, and Girshick]{gupta2019lvis}
Agrim Gupta, Piotr Doll{\'a}r, and Ross Girshick.
\newblock {LVIS}: A dataset for large vocabulary instance segmentation.
\newblock In \emph{IEEE/CVF Conference on Computer Vision and Pattern
  Recognition (CVPR)}, 2019.

\bibitem[Luo et~al.(2023)Luo, Rockwell, Lee, and Johnson]{luo2023scalable}
Tiange Luo, Chris Rockwell, Honglak Lee, and Justin Johnson.
\newblock Scalable {3D} captioning with pretrained models.
\newblock In \emph{Advances in Neural Information Processing Systems, Datasets
  and Benchmarks Track (NeurIPS D\&B)}, 2023.

\bibitem[Yang et~al.(2025)Yang, Li, Yang, Zhang, Hui, Zheng, Yu, Gao, Huang,
  Lv, et~al.]{yang2025qwen3}
An~Yang, Anfeng Li, Baosong Yang, Beichen Zhang, Binyuan Hui, Bo~Zheng, Bowen
  Yu, Chang Gao, Chengen Huang, Chenxu Lv, et~al.
\newblock {Qwen3} technical report.
\newblock \emph{arXiv preprint arXiv:2505.09388}, 2025.

\bibitem[Fischler and Bolles(1981)]{fischler1981random}
Martin~A. Fischler and Robert~C. Bolles.
\newblock Random sample consensus: A paradigm for model fitting with
  applications to image analysis and automated cartography.
\newblock \emph{Communications of the ACM (CACM)}, 24\penalty0 (6):\penalty0
  381--395, 1981.

\bibitem[Guerreiro et~al.(2023)Guerreiro, Nakazawa, and
  Stenger]{guerreiro2023PCTNet}
Julian Jorge~Andrade Guerreiro, Mitsuru Nakazawa, and Bj{\"o}rn Stenger.
\newblock {PCT-Net}: Full resolution image harmonization using pixel-wise color
  transformations.
\newblock In \emph{IEEE/CVF Conference on Computer Vision and Pattern
  Recognition (CVPR)}, 2023.

\bibitem[Radford et~al.(2021)Radford, Kim, Hallacy, Ramesh, Goh, Agarwal,
  Sastry, Askell, Mishkin, Clark, Krueger, and Sutskever]{radford2021learning}
Alec Radford, Jong~Wook Kim, Chris Hallacy, Aditya Ramesh, Gabriel Goh,
  Sandhini Agarwal, Girish Sastry, Amanda Askell, Pamela Mishkin, Jack Clark,
  Gretchen Krueger, and Ilya Sutskever.
\newblock Learning transferable visual models from natural language
  supervision.
\newblock In \emph{International Conference on Machine Learning (ICML)}, 2021.

\bibitem[Niu et~al.(2025)Niu, Hu, Lin, Jiang, and Zhang]{niu2025EOVSeg}
Hongwei Niu, Jie Hu, Jianghang Lin, Guannan Jiang, and Shengchuan Zhang.
\newblock {EOV-Seg}: Efficient open-vocabulary panoptic segmentation.
\newblock In \emph{AAAI Conference on Artificial Intelligence (AAAI)}, 2025.

\bibitem[Xiao et~al.(2025)Xiao, Kabra, Li, Lee, Carreira, and
  Panda]{xiao2025owsam}
Shiting Xiao, Rishabh Kabra, Yuhang Li, Donghyun Lee, Joao Carreira, and
  Priyadarshini Panda.
\newblock {OpenWorldSAM}: Extending {SAM2} for universal image segmentation
  with language prompts.
\newblock In \emph{Advances in Neural Information Processing Systems
  (NeurIPS)}, 2025.

\bibitem[Liu et~al.(2022)Liu, Mao, Wu, Feichtenhofer, Darrell, and
  Xie]{liu2022convnet}
Zhuang Liu, Hanzi Mao, Chao-Yuan Wu, Christoph Feichtenhofer, Trevor Darrell,
  and Saining Xie.
\newblock A {ConvNet} for the 2020s.
\newblock In \emph{IEEE/CVF Conference on Computer Vision and Pattern
  Recognition (CVPR)}, 2022.

\bibitem[Ravi et~al.(2025)Ravi, Gabeur, Hu, Hu, Ryali, Ma, Khedr, R{\"a}dle,
  Rolland, Gustafson, Mintun, Pan, Alwala, Carion, Wu, Girshick, Doll{\'a}r,
  and Feichtenhofer]{ravi2024sam2}
Nikhila Ravi, Valentin Gabeur, Yuan-Ting Hu, Ronghang Hu, Chaitanya Ryali,
  Tengyu Ma, Haitham Khedr, Roman R{\"a}dle, Chloe Rolland, Laura Gustafson,
  Eric Mintun, Junting Pan, Kalyan~Vasudev Alwala, Nicolas Carion, Chao-Yuan
  Wu, Ross Girshick, Piotr Doll{\'a}r, and Christoph Feichtenhofer.
\newblock {SAM} 2: Segment anything in images and videos.
\newblock In \emph{International Conference on Learning Representations
  (ICLR)}, 2025.

\bibitem[Mohan et~al.(2026{\natexlab{a}})Mohan, Hurtado, Mohan, and
  Valada]{mohan2026forecastocc}
Riya Mohan, Juana~Valeria Hurtado, Rohit Mohan, and Abhinav Valada.
\newblock Forecastocc: Vision-based semantic occupancy forecasting.
\newblock \emph{arXiv preprint arXiv:2602.08006}, 2026{\natexlab{a}}.

\bibitem[Hindel et~al.(2025)Hindel, Mohan, Bratuli{\'c}, Cattaneo, Brox, and
  Valada]{hindel2025label}
Julia Hindel, Rohit Mohan, Jelena Bratuli{\'c}, Daniele Cattaneo, Thomas Brox,
  and Abhinav Valada.
\newblock Label-efficient lidar semantic segmentation with 2d-3d vision
  transformer adapters.
\newblock In \emph{2025 IEEE/RSJ International Conference on Intelligent Robots
  and Systems (IROS)}, pages 4115--4122. IEEE, 2025.

\bibitem[Hurtado et~al.(2026)Hurtado, Mohan, and
  Valada]{hurtado2026hyperspectral}
Juana~Valeria Hurtado, Rohit Mohan, and Abhinav Valada.
\newblock Hyperspectral adapter for semantic segmentation with vision
  foundation models.
\newblock \emph{IEEE Robotics and Automation Letters}, 2026.

\bibitem[Mohan et~al.(2026{\natexlab{b}})Mohan, Drews, Miron, Cattaneo, and
  Valada]{mohan2026up}
Rohit Mohan, Florian Drews, Yakov Miron, Daniele Cattaneo, and Abhinav Valada.
\newblock Up-fuse: Uncertainty-guided lidar-camera fusion for 3d panoptic
  segmentation.
\newblock \emph{arXiv preprint arXiv:2602.19349}, 2026{\natexlab{b}}.

\bibitem[He et~al.(2016)He, Zhang, Ren, and Sun]{he2016deep}
Kaiming He, Xiangyu Zhang, Shaoqing Ren, and Jian Sun.
\newblock Deep residual learning for image recognition.
\newblock In \emph{IEEE Conference on Computer Vision and Pattern Recognition
  (CVPR)}, 2016.

\bibitem[Carion et~al.(2026)Carion, Gustafson, Hu, Debnath, Hu, Coll-Vinent,
  Ryali, Alwala, Khedr, Huang, Lei, Ma, Guo, Kalla, Marks, Greer, Wang, Sun,
  R{\"a}dle, Afouras, Mavroudi, Xu, Wu, Zhou, Momeni, Hazra, Ding, Vaze,
  Porcher, Li, Li, Kamath, Cheng, Doll{\'a}r, Ravi, Saenko, Zhang, and
  Feichtenhofer]{carion2026sam3}
Nicolas Carion, Laura Gustafson, Yuan-Ting Hu, Shoubhik Debnath, Ronghang Hu,
  Didac~Suris Coll-Vinent, Chaitanya Ryali, Kalyan~Vasudev Alwala, Haitham
  Khedr, Andrew Huang, Jie Lei, Tengyu Ma, Baishan Guo, Arpit Kalla, Markus
  Marks, Joseph Greer, Meng Wang, Peize Sun, Roman R{\"a}dle, Triantafyllos
  Afouras, Effrosyni Mavroudi, Katherine Xu, Tsung-Han Wu, Yu~Zhou, Liliane
  Momeni, Rishi Hazra, Shuangrui Ding, Sagar Vaze, Francois Porcher, Feng Li,
  Siyuan Li, Aishwarya Kamath, Ho~Kei Cheng, Piotr Doll{\'a}r, Nikhila Ravi,
  Kate Saenko, Pengchuan Zhang, and Christoph Feichtenhofer.
\newblock {SAM} 3: Segment anything with concepts.
\newblock In \emph{International Conference on Learning Representations
  (ICLR)}, 2026.

\end{thebibliography}
}

\end{document}